\documentclass[10pt]{article}
\usepackage[a4paper, left=20mm, right=20mm]{geometry}

\usepackage{amsthm,amsmath,amssymb}
\usepackage[numbers]{natbib}
\usepackage{mathtools}
\usepackage{algorithm}
\usepackage{algpseudocode}
\usepackage{xspace}
\usepackage{booktabs}
\usepackage{makecell}
\usepackage{listings}
\usepackage{url}
\usepackage{fnpct}
\usepackage{tikz}
\usepackage{forest}

\usepackage{pgf,pgfplots,pgfplotstable}
\usepgfplotslibrary{external}
\usetikzlibrary{positioning}
\usetikzlibrary{pgfplots.statistics,arrows,backgrounds}
\pgfplotsset{compat=1.14}

\usepackage{tabularx}
\newcolumntype{R}{>{\raggedleft\arraybackslash}X}

\newtheorem{Definition}{Definition}

\newcommand{\BE}{boundary estimation\xspace}
\newcommand{\method}{Bion\xspace}

\newcommand{\z}{\ensuremath{\mathbf{z}}\xspace}
\newcommand{\fz}{\ensuremath{f_{\z}}\xspace}
\newcommand{\zopt}{\ensuremath{\z_{opt}}\xspace}

\newcommand{\lb}[1]{\ensuremath{\underline{\mathbf{#1}}}\xspace}
\newcommand{\ub}[1]{\ensuremath{\overline{\mathbf{#1}}}\xspace}
\newcommand{\estzub}{\ensuremath{\hat{\ub{\z}}}\xspace}
\newcommand{\estzlb}{\ensuremath{\hat{\lb{\z}}}\xspace}

\newcommand{\obj}{\ensuremath{\mathbf{z}}\xspace}
\newcommand{\vars}{\ensuremath{\mathcal{V}}\xspace}
\newcommand{\constraints}{\ensuremath{\mathcal{C}}\xspace}
\newcommand{\gtba}{\ensuremath{\text{GTB}_{a}}\xspace}
\newcommand{\gtbs}{\ensuremath{\text{GTB}_{s}}\xspace}
\newcommand{\nna}{\ensuremath{\text{NN}_{a}}\xspace}
\newcommand{\nns}{\ensuremath{\text{NN}_{s}}\xspace}

\DeclareUnicodeCharacter{2212}{-}

\usepackage{comment}
\usepackage{todonotes}
\presetkeys{todonotes}{inline}{}

\title{Predictive Machine Learning of Objective Boundaries for Solving COPs}

\author{Helge Spieker \and Arnaud Gotlieb}
\date{Simula Research Laboratory, Oslo, Norway; \{helge,arnaud\}@simula.no}

\begin{document}

\maketitle

\begin{abstract}Solving Constraint Optimization Problems (COPs) can be dramatically simplified by boundary estimation, that is, providing tight boundaries of cost functions. By feeding a supervised Machine Learning (ML) model with data composed of known boundaries and extracted features of COPs, it is possible to train the model to estimate boundaries of a new COP instance. In this paper, we first give an overview of the existing body of knowledge on ML for Constraint Programming (CP) which learns from problem instances. Second, we introduce a boundary estimation framework that is applied as a tool to support a CP solver. Within this framework, different ML models are discussed and evaluated regarding their suitability for boundary estimation, and countermeasures to avoid unfeasible estimations that avoid the solver to find an optimal solution are shown. Third, we present an experimental study with distinct CP solvers on seven COPs. Our results show that near-optimal boundaries can be learned for these COPs with only little overhead. These estimated boundaries reduce the objective domain size by 60-88\% and can help the solver to find near-optimal solutions early during search.\end{abstract}

\section{Introduction}
Constraint Optimization Problems (COPs) are commonly solved by systematic
tree search, such as branch-and-bound, where a specialized solver prunes those parts of the search space
with worse cost than the current best solution. In Constraint
Programming (CP), these systematic techniques work without prior knowledge and
are steered by the constraint model. However, the worst-case computational cost
to fully explore the search space is exponential and the search performance depends on solver configuration, such as selection of right parameters, heuristics, and search strategies, as well as appropriate formulations of the constraint problems to enable efficient pruning of the search space.

Machine Learning (ML) methods, on the other hand, are data-driven and
are trained with labeled data or by interaction with their environment, without explicitly considering the problem
structure or any solving procedure. At the same time, ML methods can only approximate the
optimum and are therefore not a full alternative. The main computational cost of
these methods lies in their training, but the cost when estimating an outcome
for a new input is low. This low inference cost makes ML an interesting candidate for
integration with more costly tree search.
Previous research has examined this integration to several extents, albeit selecting the appropriate algorithm within a solver and to configure it for a given instance, learning additional constraints to model a problem, or learning partial solutions.

\begin{figure}[t]
    \centering
    \includegraphics[width=\columnwidth]{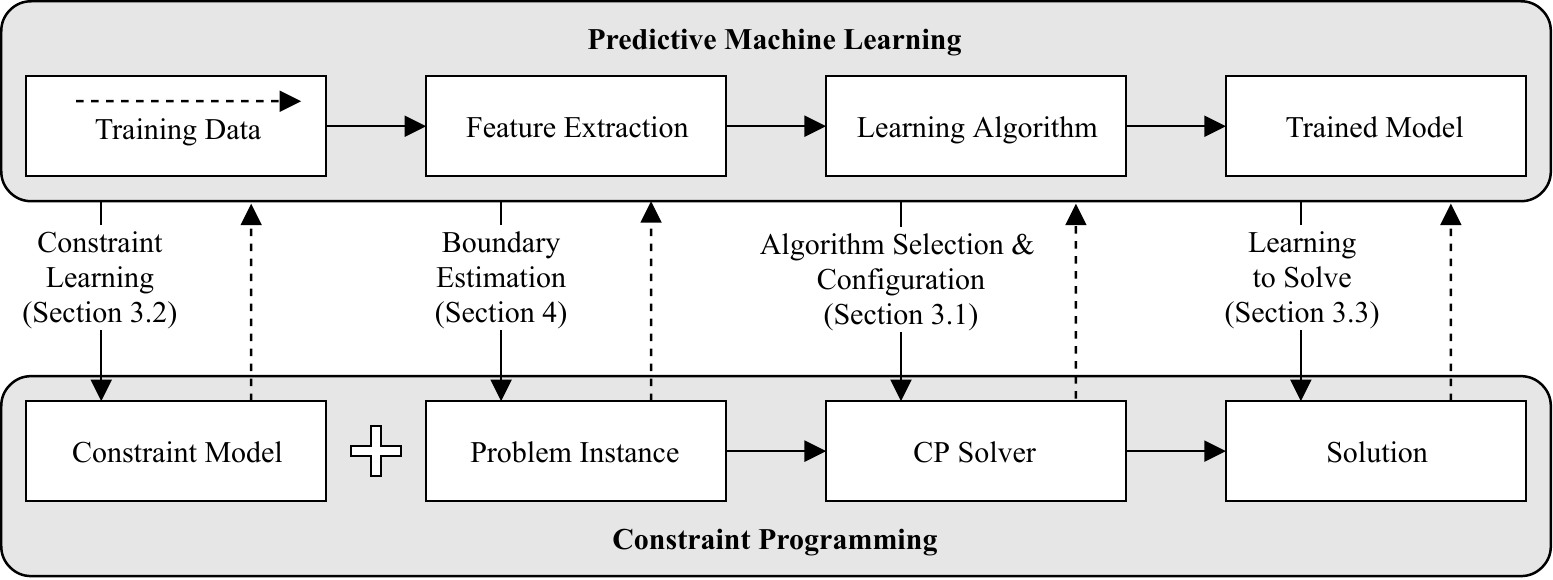}
    \caption{Overview of the applications of ML and CP considered in this paper. Depending on the application type, training data can be extracted at different points in the CP solving process.}
    \label{fig:overview}
\end{figure}

In this paper, we provide an overview of the body of knowledge on using ML for CP. 
We focus on approaches where the ML model is trained in a supervised way from existing problem instances with solutions, and collected information gathered while solving them.
We discuss of the characteristics of supervised ML and CP and how to use predictive ML to improve CP solving procedures.
A graphical overview of predictive ML applications for CP is shown in Figure~\ref{fig:overview}.

The second part of the paper discusses a boundary estimation method called \method, which combines logic-driven constraint
optimization and data-driven ML, for solving COPs. A ML-based estimation model
predicts boundaries for the objective variable of a problem instance, which can
then be exploited by a CP solver to prune the search space. 
To reduce the risk of inaccurate estimations, which can render the COP unsatisfiable, the ML model is trained
with an asymmetric loss function and adjusted training labels. 
Setting close bounds for the objective variable is helpful to prune the search space \cite{Milano2006,Gualandi2012}. 
However, estimating tight bounds in a problem- and solver-agnostic way is still an open problem \cite{Ha2015}. 
Such a generic method would allow many COP instances to be solved more efficiently.
Our method has first been introduced in \cite{Spieker2020a}, where an initial analysis of the results led us to conclude that \method was a promising approach to solve COPs. 
Here, we revisit and broaden the presentation of \method in the general context of predictive machine learning for constraint optimization.
We extend our experimental evaluation analysis and show that boundary estimation is effective to support COP solving.

\method is a two-phase procedure: First, an objective boundary for
the problem instance is estimated with a previously trained ML regression model~; 
Second, the optimization model is extended by a boundary constraint on the objective
variable and a constraint solver is used to solve the problem instance. This
two-phases approach decouples the boundary estimation from the actual solver and
enables its combination with different optimized solvers. Training the estimation
model with many problem instances is possible without any domain- or expert- knowledge and
applies to a wide range of problems.
Solving practical assignment, planning or scheduling problems often requires to
repeatedly solve the very same COP with different inputs. \method is
well fitted for these problems, where training samples, which resemble realistic
scenarios, can be collected from previous iterations. Besides, for any COP,
\method can be used to pre-train problem-specific ML models which are then
deployed to solve new instances.

The remainder of this paper is structured as follows.
In Section \ref{sec:background}, we discuss the necessary background for the integration of predictive ML and CP, 
including data curation and preparation of problem instances to be usable in ML. 
Section \ref{sec:predml} reviews existing work on predictive ML and CP for algorithm selection and configuration, constraint learning, and learning to solve.
Afterwards, we introduce \method in Section \ref{sec:boundaryestimation}, a ML-based method to estimate boundaries of the objective variable of COPs as a problem- and solver-independent method tool to support constraint optimization solvers. 
We discuss training techniques to avoid inaccurate estimations, compare various ML models such as gradient tree boosting, support vector machine and neural networks, with symmetric and asymmetric loss functions. 
A technical CP contribution lies in the dedicated feature selection and user-parameterized label shift, a new method to train these models on COP characteristics.
In Section \ref{sec:experiments}, \method's ability to prune objective domains, 
as well as the impact of the estimated boundaries on solver performance, is evaluated on seven COPs that were previously used in the MiniZinc challenges to compare CP solvers efficiency.
Finally, Section \ref{sec:conclusion} concludes the paper with an outlook on future applications of predictive ML to CP.

\section{Background} \label{sec:background}
This section presents a background discussion on constraint optimization and supervised machine learning.
We introduce the necessary terminology and foundations under the context of predictive ML for CP.
For an in-depth introduction beyond the scope of this paper, we refer the interested reader to the relevant literature in constraint optimization \cite{Rossi2006,Marriott1998} and machine learning \cite{Hastie2009,Domingos2012,Murphy2022}.

\subsection{Constraint Optimization Problems}

Throughout this paper, constraint optimization is considered in the context of Constraint Programming
 over Finite Domains \cite{Rossi2006}. In that respect,
every variable $\mathbf{x}$ of an optimization problem is associated with a finite set of 
possible values, called a Finite Domain (FD) and, noted $D_\mathbf{x}$. 
Each value is associated with a unique integer without any
loss of generality and thus, $D_\mathbf{x} \subseteq \underline{\mathbf{x}}..\overline{\mathbf{x}}$,
where $\underline{\mathbf{x}}$ (resp. $\overline{\mathbf{x}}$) denotes the lower
(resp. upper) bound of $D_\mathbf{x}$.
Each variable $\mathbf{x}$ takes one, yet unknown value in its domain, i.e., $\mathbf{x} \in D_\mathbf{x}$ and $\underline{\mathbf{x}} \leq \mathbf{x} \leq \overline{\mathbf{x}}$, even if not all the values of $\underline{\mathbf{x}}..\overline{\mathbf{x}}$ are necessarily part of $D_{\mathbf{x}}$.

A constraint is a relation among a subset of the decision variables, which restrain 
the possible set of tuples of values taken by these variables. In Constraint Programming over FD, different
type of constraints can be considered, including arithmetical, logical, symbolic or global relations. 
For instance, $\mathbf{x} + \mathbf{y} * \mathbf{z} = 5$ is an arithmetical constraint while 
$\mathbf{x=f}\, \lor\, \mathbf{y=t}\, \lor\, (\mathbf{z} > 0) = \mathbf{t}$, where $\mathbf{t}$ (resp. $\mathbf{f}$) stands for True (resp. False), is a logical constraint. Symbolic and global constraints 
include a large panel of non-fixed length relations such as \texttt{all\_different}$([\mathbf{x_1,\dots,x_n}])$, which 
constrains each variable $\mathbf{x_i}$ to take a different value, \texttt{element}$([\mathbf{x_1,\dots,x_n}], \mathbf{i,v})$, which
enforces the relation $\mathbf{v}=\mathbf{x_i}$, where $\mathbf{v,i}$ and $\mathbf{x_1,\dots,x_n}$ can all be unknowns. 
More details and examples can be found in \cite{Rossi2006,Marriott1998}.

\begin{Definition}[Constrained Optimization Problem (COP)]
  A COP is a triple $\langle \vars, \constraints, f_\obj \rangle$ where \vars
  denotes a set of FD variables, called {\it decision variables}, $\constraints$ denotes a set of
  constraints and $f_\obj$ denotes an optimization function which depends on the
  variables of \vars. $f_\obj$'s value ranges in the domain of the
  variable \obj %
  (\obj in $\underline{\mathbf{z}}..\overline{\mathbf{z}}$), called the {\it objective variable}.
\end{Definition}
Solving a COP instance requires finding a variable assignment, 
i.e., the assignment of each decision variable to a unique value from its domain, 
such that all constraints are satisfied and the objective variable \obj takes an optimal value. Note that COPs have to be distinguished from well-known {\it Constraint Satisfaction Problems} (CSPs) where the goal is only to find satisfying variable assignments, without taking care of the objective variable.

\begin{Definition}[Feasible/Optimal Solutions]
  Given a COP instance $\langle \vars, \constraints, f_\obj \rangle$, a {\it feasible solution} 
  is an assignment of all variables \vars in their domain, 
  such that all constraints are satisfied.
  $\obj_{cur}$ denotes the value of objective variable \obj for such a feasible
  solution. An {\it optimal solution} is a feasible solution, which optimizes the
  function $f_\obj$ to the optimal objective value $\obj_{opt}$.
\end{Definition}

\begin{Definition}[Satisfiable/Unsatisfiable/Solved COP]
  A COP instance $\langle \mathcal{V}, \mathcal{C}, f_\obj \rangle$ is
  {\it satisfiable} (resp. {\it unsatisfiable}) iff it has at least one feasible solution
  (resp. no solution). A COP is said to be {\it solved} if and only if at least one of its  
  optimal solutions is provided.
\end{Definition}

Solving a COP instance can be done by a typical \textit{branch-and-bound} search process
which incrementally improves a feasible solution until an optimal solution is
found. Roughly speaking, in case of minimization, branch-and-bound works as
follows: starting from an initial feasible solution, it incrementally adds to
$\mathcal{C}$ the constraint $\obj < \obj_{cur}$, such that any later found
feasible solution has necessarily a smaller objective value than the current value.
This is helpful to cut the search tree of all feasible solutions which have a
value equal to or larger than the current one. If there is no smaller feasible
value and all possible variable assignments have been explored, then the current
solution is actually an optimal solution.
Interestingly, the search process can be time-controlled and interrupted at any time. 
Whenever the search process is interrupted before completion, it returns
$\obj_{cur}$ as the best feasible solution found so far by the search process, i.e., 
a near-optimal solution. This solution is provided with neither proof of
optimality nor guarantee of proximity to an optimal solution, but it is still 
sufficient in many applications.

This branch-and-bound search process is fully implemented in many Constraint
Programming (CP) solvers. These solvers provide a wide range of features and
heuristics to tune the search process for specific optimization problems. At the
same time, they do not reuse any of the already-solved instances of a constraint problem to
improve the optimization process for a new instance. In addition to presenting existing works on predictive 
learning for constraint optimization, this paper proposes a new method to
reuse existing known boundaries to nurture a machine learning model to improve
the optimization process for a new instance.

\subsection{Supervised Machine Learning} \label{sec:mlmodels}
At the core of the predictive applications described in this paper, a supervised ML model is trained and deployed.
In supervised machine learning, the model is trained from labeled input/output examples to approximate the underlying, usually unknown function.

\begin{Definition}[Supervised Machine Learning]
Given a set of training examples $\{(x_1,y_1),\allowbreak(x_2,y_2),\allowbreak\dots,\allowbreak(x_n,y_n)\}$, a supervised machine learning model approximates a function $P: X \rightarrow Y$, with $X$ being the input space, and $Y$ being the output space.
Here, $x_i$ is a vector of instance features and $y_i$ is the corresponding label, representing the target value to be predicted.
\end{Definition}

Every value in $x$ corresponds to a \textit{feature}, that is, a problem instance characteristic that describes the input to the model. In most cases, the input consists of multiple features and is also described as the \textit{feature vector}.
The output of the model, $\hat{y}$, is defined as a vector, too, although it is more common to have a model that only predicts a single value. This is the case in regression problems, when predicting a continuous value, or binary classification, when deciding whether to activate a functionality or not. %

The model $P$ is trained from a training set, consisting of example instances $(x_i,y_i)$.
Training the model describes the process to minimize the error between estimated value $\hat{y}$ and true (observed) value $y$ of the training examples.
The error is assessed with a loss function, that can be different depending on the task of the machine learning model.
We show examples for two commonly used types of loss functions.
For regression problems, where a continuous output value is estimated, the loss is assessed via the \textit{mean squared error}.
In classification, where the input is assigned to one of multiple classes, the \textit{cross-entropy loss} is calculated.

\begin{Definition}[Mean Squared Error (MSE)]
Given a set of $N$ estimated and observed target values $\{(\hat{y}_1, y_1),(\hat{y}_2, y_2),\dots,(\hat{y}_N, y_N)\}$,
the MSE is calculated as: 
\[L = \frac{1}{N} \sum_{i=1}^{N} (\hat{y}_i - y_i)^2\]

Within the calculation of MSE, positive and negative errors have the same effect, but larger errors are stronger penalized, i.e. they have a larger influence, than smaller errors.
\end{Definition}

\begin{Definition}[Cross-Entropy Loss]
Given a set of $N$ estimates and $K$ classes, the cross-entropy (also log-likelihood) is calculated as:
\[L = -\frac{1}{N} \sum_{i=1}^{N} \sum_{k=1}^{K} y_{ik} \log \hat{y_{ik}}\]
with $\hat{y_{ik}}$ being the probability of $x_i$ belonging to class $k$ and 
\[
y_{ik} =
\begin{cases}
1 & \text{if $x_i$ belongs to class $k$}\\
0 & \text{otherwise}
\end{cases}
\]
\end{Definition}

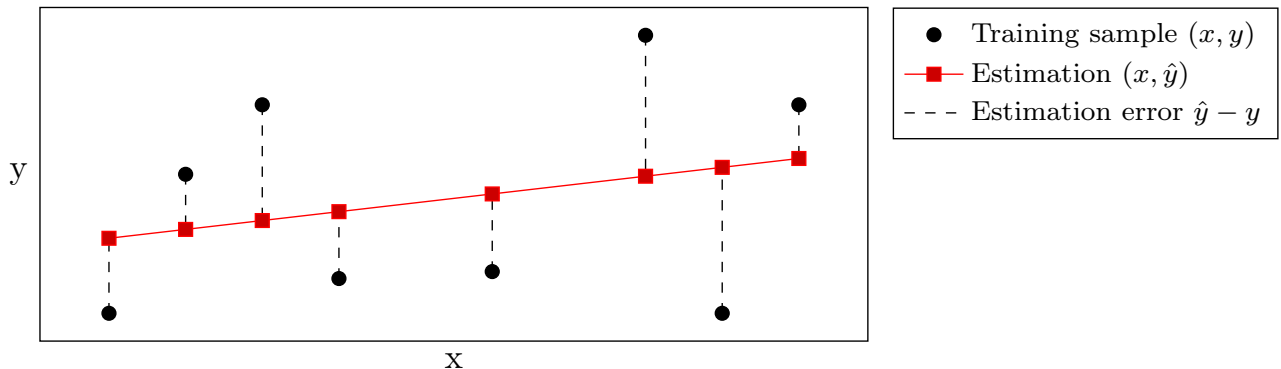
\begin{figure}[t]
    \centering
    \resizebox{\columnwidth}{!}{%
\begin{tikzpicture}
    \pgfplotsset{
        width=10cm,
        height=5cm,
        compat=1.4,
        legend style={font=\footnotesize}}
    \begin{axis}[
    xlabel={x},
    ylabel={y},
    ylabel style={rotate=-90},
    legend cell align=left,
    legend pos=outer north east,
    xticklabels={,,},
    yticklabels={,,},
    ticks=none]
    \addplot[only marks] table[row sep=\\]{
        X Y\\
        1 0.5\\
        2 1.5\\
        3 2\\
        4 0.75\\
        6 0.8\\
        8 2.5\\
        9 0.5\\
        10 2\\
    };
    \addlegendentry{Training sample $(x,y)$}
    \addplot table[row sep=\\,
    y={create col/linear regression={y=Y}}] %
    {
        X Y\\
        1 0.5\\
        2 1.5\\
        3 2\\
        4 0.75\\
        6 0.8\\
        8 2.5\\
        9 0.5\\
        10 2\\
    };
    \addlegendentry{Estimation $(x,\hat{y})$}
    \addplot[mark=none, dashed] coordinates {(1,0.5) (1,1.04)};
    \addplot[mark=none, dashed] coordinates {(2,1.5) (2,1.11)};
    \addplot[mark=none, dashed] coordinates {(3,2)    (3,1.17)};
    \addplot[mark=none, dashed] coordinates {(4,0.75) (4,1.24)};
    \addplot[mark=none, dashed] coordinates {(6,0.8) (6,1.36)};
    \addplot[mark=none, dashed] coordinates {(8,2.5) (8,1.49)};
    \addplot[mark=none, dashed] coordinates {(9,0.5) (9,1.55)};
    \addplot[mark=none, dashed] coordinates {(10,2)  (10,1.62)};
    \addlegendentry{Estimation error $\hat{y} - y$}
    \end{axis}
\end{tikzpicture}
}
    \caption{Training a supervised machine learning model: Illustrative example for linear regression $y = a_0 + a_1*x$. During training the weights $a_0, a_1$ are adjusted to minimize the estimation error.}
    \label{fig:mltraining}
\end{figure}

An example for the training scheme is shown in Figure \ref{fig:mltraining} for a linear regression model.
The weights $a_0, a_1$ of the linear function are adjusted such that the total error between estimated and true values is minimized.
In the given example the training examples do not strictly follow a linear trend and are therefore difficult to approximate with only a small error.
This is an indicator to use a more complex model for better results and to describe the output via different features than only $x$, if possible.

\subsection{Machine Learning Models}
Many supervised ML models exist and have been shown to be applicable to a wide range of problems. 
However, there is no one best model and depending on the application, different models can show good performance.

In this section, we introduce five widely used machine learning models for the
application in predictive ML for CP: Gradient Tree Boosting, Neural Network,
Support Vector Machine, k-Nearest Neighbors, and Linear Regression.
We briefly discuss each model and highlight relevant
characteristics for applying them for constraint problems.

\subsubsection{Gradient Tree Boosting}
Gradient Tree Boosting (GTB), also Gradient Boosting Machine, 
is an ensemble method where multiple individual weak models, whose error rate slightly outperforms random guessing, are combined into a strong learner \cite[]{Friedman2001}.
At each iteration, additional weak models are trained on
a modified subset of data to add information to the previous prediction.
The individual weak models in GTB are decision trees, which by themselves have weak estimation accuracy compared to other models, but are robust to handle different types of inputs and features \cite[Ch. 10]{Hastie2009}.

\subsubsection{Neural Network}
A multi-layer Neural Network (NN) approximates the function to-be-learned
over multiple layers of nodes or neurons. Neural networks can be applied for
different problems, e.g. classification or regression, especially when there is
a large amount of training data.
Designing a neural network requires selecting an architecture and the number of layers and nodes, as well as performing a careful hyper-parameter optimization to achieve accurate results \cite[]{Domingos2012}.

An ensemble of multiple NNs can be formed to reduce the
generalization error of a single NN \cite[]{Zhou2002}, by taking the average of all
predictions. As errors are expected to be randomly distributed around the actual
value, an ensemble can reduce the error.

\subsubsection{Support Vector Machine}
Support Vector Machines (SVM) map their inputs into a high-dimensional feature
space. This feature space is defined by a kernel function, which
is a central component. Once the data has been mapped, linear
regression is performed in this high-dimensional space \cite[]{Cortes1995}.
One common variant for regression problems are \(\epsilon\)-SVR, which are usually trained using a soft
margin loss \cite[]{Smola2004}. Under soft margin loss an error is penalized only
if it is larger than a parameter \(\epsilon\), otherwise it is similar to
a squared error function. Having the additional margin of allowed errors avoids minimal adjustments during training and allows for higher robustness of the final model.

\subsubsection{Nearest Neighbors}
Nearest neighbors methods, also k-Nearest Neighbors (kNN) or neighbors, relate
an unseen instance to the closest samples, i.e. the nearest neighbors, in the
training set \cite[]{Larose2004}. The distance between points is calculated from a
distance metric, which is often the euclidean distance. In case of k-nearest
neighbors, the number of neighbors to consider is fixed to $k$, which is usually
a small integer value. Other methods set the number of neighbors dynamically
from the data density in the training set and a threshold for the maximum
distance. For regression problems, the estimated value $y$ is calculated by a
weighted average over the neighbors' values. The weights are either uniform or
proportional to the distance.

Nearest neighbor methods have the advantage to be simple and non-parameterized,
i.e. they do not require a training phase, but the complete training is
necessary to process new instances. Because searching through all training
samples for each estimation is inefficient for a large training set, tree-based
data structures can be used to organize the data for faster access, for example,
K-D trees \cite[]{Bentley1975} or Ball trees \cite[]{Omohundro1989}.

\subsubsection{Linear Regression}
Linear regression (LR) is a simple statistical approach to find a linear relationship between a set of input features and the target value.
Applying linear regression is effective in scenarios where a linear relationship can be assumed.
In other scenarios, LR is less accurate than the other introduced methods.
However, because it is easy to train and apply, LR is commonly used as a baseline method to identify and justify the need for more complex, non-linear methods in ML applications.

The model is formed by the linear relationships between each of the \(n\) features and the
target value, the dependent variable \(y\). 
This relationship is captured by the parameter \(a_i\) for each feature \(x_i\):
\(y = a_0 + a_1 x_1 + a_2 x_2 + \dots + a_n x_n\). 
Linear regression is trained via the \textit{ordinary least squares} method \cite[Ch. 3]{Hastie2009}, an iterative method to minimize the squared error between estimated and true target.

\subsection{Data Curation}

Machine learning methods are data-driven and need a data corpus to be trained,
before they can be used to make estimations on new instances. In this section,
we discuss the collection of a data corpus, its organization for training the
model, and the pre-processing to transform the data into a format that is usable as model input.

\subsubsection{Collection}

A sufficient amount of training data is the basis to train a supervised ML model.
Data for CSPs and COPs, that is, problem instances, can either be downloaded
from open repositories for existing constraint problems, collected from
historical data, or synthetically generated.

For many problems, constraint models and instances can be freely accessed from online repositories.
The CSP library (CSPLib) \cite[]{Jefferson1999} contains a large collection of constraint models, instances, and their results in different modeling languages.
Furthermore, problem-specific libraries exist, such as TSPLib \cite[]{Reinelt1991} for the traveling sales person problem and related problems, or ASLib \cite[]{Bischl2016} for algorithm selection benchmarks.
Finally, there are repositories of constraint models and instances in language-specific repositories, for example in MiniZinc\footnote{Online at: \url{https://github.com/MiniZinc/minizinc-benchmarks}} or XCSP3\footnote{Online at: \url{http://www.xcsp.org/}}.

Having a generator allows us to create a large training corpus for a particular problem, 
but as it also requires additional effort to develop the generator program. This solution might not be suitable in all cases. 
Another approach is to generate instances directly from the constraint model \cite[]{Gent2014}.
The constraint model is reformulated by defining the given instance parameters as variables to be found
by the solver. Solving this reformulation with random value assignment then
leads to a satisfiable problem instance of the original constraint model.

However, for all generators, the difference between generated instances, that
are distributed over the whole possible instance space, and realistic
instances, that might only occupy a small niche of the possible instance space,
has to be considered by either adjusting the generator to create realistic
instances or to ensure the training and test set include realistic instances
from other sources.

\subsubsection{Data Organization}

Data used to build a ML model is split into three parts: a) the training
set, b) the development (dev) or validation set, and c) the test set. 
The training set is used to train the model via a learning algorithm, whereas the
dev set is only used to control the parameters of the learning algorithm. 
The test set is not used during training or to adjust any parameters, but only
serves to evaluate the performance of the trained model. Especially the test set
should be similar to those instances that are most likely to be encountered in
practical applications.

The training set holds the largest part of the data, ranging from 50-80\% of the data, while the rest of the data is equally divided between validation and test set. 
This is a rough estimate and the exact split is dependent on the total size of the data set. 
In any case, it should be ensured the validation and test set are sufficiently large to evaluate the trained model.

\subsubsection{Representation}

The representation refers to the format into which a problem instance is transformed before it can be used as ML input.
An expressive representation is crucial for the design of a ML model with high influence on its later performance.
Representation consists of feature selection and data preparation, which are introduced in the following.

\paragraph{Feature Selection}
Feature selection defines which information is available for the model to make predictions and if insufficient or the wrong information is present, it is not possible to learn an accurate prediction model, independent of the selected machine learning technique.
A good feature selection contains all features which are necessary to calculate the output and captures relations between instance data and the quantity to estimate.

As a long-term vision, it is desirable to learn a model end-to-end, that is from the raw COP formulation and instance parameters, without having to extract handcrafted features.
Currently, most machine learning techniques work with fixed, numerical input and output vectors.
There are machine learning techniques capable to handle variable-length inputs and outputs, for example, recurrent neural networks like LSTM \cite[]{Hochreiter1997}, but these have, to the best of our knowledge, not yet been successfully applied in the area of predictive ML for CP and will not be further discussed here.
Instead we focus on the common case to handcraft a fixed-size feature vector.

For the selection of features, we first need to consider the application of the machine learning model.
Is it a problem-specific application, that handles only instances of one defined optimization problem, or is it a problem-independent application, that handles instances of many different optimization problems?

In COP-specific applications, domain knowledge can be exploited. For example, when building a predictive model for the travelling sales person (TSP) problem \cite[]{Hutter2014}, features describing the spatial distribution of the cities and the total area size are valuable \cite[]{Smith-Miles2010}. 
Similarly, the constrained vehicle routing problem (CVRP) has been investigated to identify problem-specific features that are beneficial to reason over solution quality \cite[]{Arnold2019b} or aid the search process \cite[]{Arnold2019,Accorsi2020,Lucas2020}.

Without domain knowledge, more generic features have to be used to capture the characteristics and variance of different constraint models and their instances.
One approach is the design of portfolio solvers, where a learning model is used to decide which solver to run for a given problem instance \cite[]{Xu2007,OMahony2008,Malitsky2012, Seipp2015, Amadini2015, Amadini2016a}.
Feature extraction exploits the structure of the general constraint model and the specific instance, its constraints, variables and their domains.
Features are further categorized as static features, which are constant for one model and instance, and dynamic features, which change during search and are therefore especially relevant for algorithm configuration and selection tasks.
As a representative explanation, Table \ref{tab:features} shows an overview of features to describe problem instances.
While many of these features are constant for multiple instances of the same constraint problem, e.g. the number of constants or which constraints were defined, the variables and their domains depend on the instance parameters and can offer descriptive information that discriminate several instances of the same problem.

\begin{table}[thbp]
    \centering
    \small
    \caption{Examples for static COP features from the feature extractor \texttt{mzn2feat} \cite{Amadini2014}, which analyses COPs formulated in the MiniZinc constraint modeling language \cite{Nethercote2007}. The descriptions are quoted from \cite{Amadini2014}. NV: Number of variables, NC: Number of constraints, CV: variation coefficient, H: entropy of a set of values\label{tab:features}}
    \begin{tabularx}{\columnwidth}{lX}
        \toprule
        Category & Features \\
        \midrule
        Variables & 
        The number of variables $NV$; 
        the number $cv$ of constants; 
        the number $av$ of aliases; 
        the ratio $\frac{av+cv}{NV}$;
        the ratio $\frac{NV}{NC}$; 
        the number of defined variables (i.e. defined as a function of other variables); 
        the number of introduced variables (i.e. auxiliary variables introduced during the FlatZinc conversion); 
        sum, min, max, avg, CV, and H of the: variables domain size, variables degree, domain size to degree ratio
        \\
        
        Domains & 
        The number of: 
        boolean variables $bv$ and the ratio $\frac{bv}{NV}$; 
        float variables $fv$ and the ratio $\frac{fv}{NV}$;
        integer variables $iv$ and the ratio $\frac{iv}{NV}$;
        set variables $sv$ and the ratio $\frac{sv}{NV}$;
        array constraints $ac$ and the ratio $\frac{av}{NV}$; 
        boolean constraints $bc$ and the ratio $\frac{bc}{NC}$;
        int constraints $ic$ and the ratio $\frac{ic}{NC}$;
        float constraints $fc$ and the ratio $\frac{fc}{NC}$;
        set constraints $sc$ and the ratio $\frac{sc}{NC}$;
        \\

        Constraints & 
        The total number of constraints $NC$,
        the ratio $\frac{NC}{NV}$,
        the number of constraints with FlatZinc annotations; 
        the logarithm of the product of the: constraints domain (product of the domain size of each variable in that constraint) and constraints degree; 
        sum, min, max, avg, CV, and H of the: constraints domain, constraints degree, domain to degree ratio
        \\
         
        Global Constraints & 
        The total number $gc$ of global constraints, 
        the ratio $\frac{gc}{NC}$ and the number of global constraints for each one of the 27 equivalence classes in which we have grouped the 47 global constraints 
        \\
         
        Graphs & 
        From the Constraint Graph $CG$ and the Variable Graph $VG$ we compute min, max, avg, CV, and H of the: 
        $CG$ nodes degree, 
        $CG$ nodes clustering coefficient, 
        $VG$ nodes degree, 
        $VG$ nodes diameter
        \\
        
        Solving & 
        The number of labeled variables (i.e. the variables to be assigned); 
        the solve goal; 
        the number of search annotations; 
        the number of variable choice heuristics; 
        the number of value choice heuristics
        \\
         
        Objective & 
        The domain $dom$, 
        the degree $deg$, 
        the ratios $\frac{dom}{deg}$ and $\frac{deg}{NC}$ of the variable $v$ that has to be optimized; 
        the degree $de$ of $v$ in the variable graph, 
        its diameter $di$,
        $\frac{de}{di}$,
        $\frac{di}{de}$.
        Moreover, named $\mu_{dom}$ and $\sigma_{dom}$ the mean and the standard deviation of the variables domain size and $\mu_{deg}$ and $\sigma_{deg}$ the mean and the standard deviation of the variables degree, 
        we compute 
        $\frac{dom}{\mu_{dom}}$,
        $\frac{deg}{\mu_{deg}}$,
        $\frac{dom-\mu_{dom}}{\sigma_{dom}}$,
        and $\frac{deg - \mu_{deg}}{\sigma_{deg}}$
        \\
        \bottomrule
    \end{tabularx}
\end{table} 
Several studies have been performed to analyze the ability of these generic features to characterize and discriminate COP characteristics \cite[]{Roberts2009,Hutter2013,Amadini2015a,Bischl2016}.
Their main conclusion is that a small number of features can be sufficient discriminators, but that there is no single set of best features for all constraint problems used in their experiments.
An approach to overcome this issue is therefore to start with a larger set of features than practically necessary and, perform {\it dimensionality reduction} (see the next Section for a detailed explanation) %
to remove features with little descriptive information.

\paragraph{Data Preparation} \label{sec:datapreparation}

Once the features are selected and retrieved, the next step is to pre-process the data, such that it can be used by the machine learning model, by performing dimensionality reduction and scaling.

A dimensionality reduction step can shrink the size of the feature vector.
Reducing the dimensionality, which means having less model inputs, can thereby also reduce the model complexity.
Features that are constant for all instances are removed, as well as features that only show minimal variance below a given threshold.
Other dimensionality reduction techniques, e.g. principal component analysis (PCA), apply statistical procedures to reduce the data to a lower-dimensional representation while preserving its variance. 
These reduction techniques can further reduce the number of features, but at the downside that it is no longer possible to directly interpret the meaning of each feature.

Feature scaling is necessary for many models and means to transform the values of each feature, which might in different ranges, into one common range. Scaled features reduce model complexity as it is not necessary to have the weights of a model account for different input ranges.
One common technique is to scale the feature by subtracting its mean and dividing by the standard deviation, which transforms the features to approximately resemble a normal distribution with zero mean and unit variance.
Another technique is called \textit{minmax-normalization} and scales the feature based on the smallest and largest occurring values, such that all values scaled into the range $[-1, 1]$.

Note that it is important to keep track of how each preparation step is performed on the training set, as it has to be repeated in the same way on each new instance during testing and production. 
This means the feature vector of a new instance contains the same features and each feature is scaled by the same parameters, e.g. it is scaled by the training set's mean and standard deviation.

\section{Predictive Machine Learning for Constraint Optimization}
\label{sec:predml}

Opportunities for integration of predictive ML in constraint programming are vast and relevant in several research directions \cite[]{Bengio2018}.
We first look at general categorizations and approaches to the combination of predictive ML and CP, before we discuss the body of knowledge in specialized research areas.

\begin{figure}[t]
\centering
\includegraphics[width=0.75\textwidth]{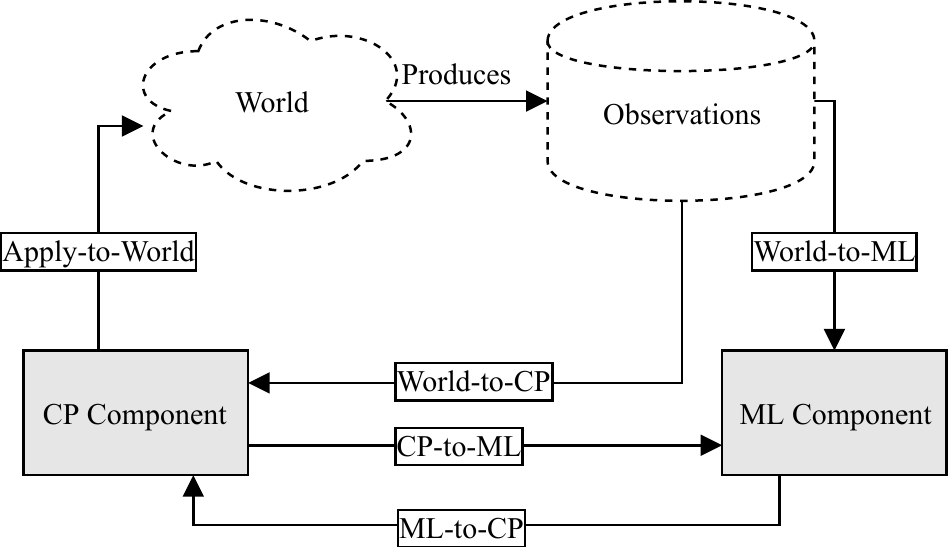}
\caption{The inductive constraint programming loop (adapted from \cite[]{Bessiere2017}). The CP and ML components can interact with each other and react upon influences from the external world and observations.}
\label{fig:icp}
\end{figure}

A recent work by \citeauthor{Bessiere2017} introduces a general framework for the integration of ML and CP, called the \textit{inductive constraint programming loop} (ICP) \cite[]{Bessiere2017}.
The framework is based on four main building blocks: a CP component, a ML component, which can be controlled, as well as an external world, that cannot be controlled, and which produces observations.
All these building blocks are interconnected and can receive and provide information from and to each other, e.g. the CP component receives new constraint problems via a World-to-CP relation and returns solutions via a corresponding Apply-to-World relation.
An overview of the ICP framework and all defined relations is shown in Figure \ref{fig:icp}.
The majority of predictive machine learning applications for CP can be embedded into the ICP framework, as they exploit the ML-to-CP relation, where the ML model transfers information to the component, which is their main purpose.
Furthermore, these applications also exploit the opposite CP-to-ML relation to return feedback from the solver, e.g. runtimes, found solutions, to the ML model for improvement.

\citeauthor{Lombardi2017} present a general framework for embedding ML-trained models in optimization techniques,
called empirical model learning \cite[]{Lombardi2017}. The approach deploys
trained ML models directly in the COP as additional global constraints.
Experimental results show, that the embedded empirical ML model can improve the total solving process.
The proposed integration further leads to easier deployment of the trained ML model and to reduce the complexity of the setup, but, at the same time, the complexity of the model itself is increased as compared to the pure COP.

\subsection{Algorithm Selection and Configuration}
The area of algorithm selection and configuration applies predictive ML to analyze individual problem instances and decide for the most appropriate solver, its heuristic, or the setting of certain tuning parameters of a solver.
All these techniques have in common, that they work on a knowledge base that is mostly not restricted to a single constraint optimization problem, but applicable to instances from many different problems.
Furthermore, these approaches affect changes onto the solver, but do not modify or adjust the constraint problem or the problem instance.

Algorithm selection within a constraint solver can be used to decide, which
search strategy to use. A search strategy consists of a variable selection, that
is which variable will next have a value assigned, and a value selection, that is which
value is assigned to the variable.
\citeauthor{Arbelaez2009} propose a classification model to select from up to 8
different heuristics, consisting of both variable and value selection \cite[]{Arbelaez2009}. 
The search strategy is repeatedly selected during search, e.g. upon
backtracking, to be able to adapt to characteristics of the problem instance in
different regions of the search space. 
The machine learning model uses a SVM (Support Vector Machine, see Section \ref{sec:mlmodels}) and
a set of 57 features to describe an instance.

In \cite[]{Arbelaez2010}, this work is extended to a life-long learning constraint
solver, whose inner machine learning model is repeatedly re-trained based on
newly encountered problem instances and the experiences from selected search heuristics.
The methodology is refined to select a heuristic for a predefined checkpoint
window, i.e. a heuristic is fixed for a sequence of decisions, before the next
heuristic is selected. In total 95 features, describing static features, such as
the problem definition, and variable and constraint information, and dynamic
features to monitor the search performance.

Similarly, \citeauthor{Gent2010} classify problem instances to decide whether solving them benefits from lazy learning, an effective but costly CSP search method \cite[]{Gent2010}. 
They analyze the primal graph of the instance to extract instance features.
The primal graph represents every variable as a node, and variables that occur in the scope of a constraint are connected via edges.
Using this graph structure allows to extract features like the edge density, graph width, or the proportion of constraints that share the same variable.

\citeauthor{Chu2015} investigate methods to learn a value selection heuristic \cite[]{Chu2015}. 
As part of their work, they discuss the problem of gathering samples to train the ML model.
Ideally, one would require exactly solved instances, but in practice this incurs high computational cost for every training instance. Their approach is to define an alternative scoring function to be used as the training target. 
This scoring function is chosen such that it does not require exact solving of the instance and therefore gathering the training data is cheaper overall.

Besides supervised ML techniques, adaptive ML methods, such as reinforcement learning, are also applied to configure search algorithms and their parameters, e.g. in large neighborhood search \cite[]{Mairy2011} or to select tree search heuristics \cite[]{Loth2013}.

Reliable information about expected runtimes for an
algorithm on a problem instance can be helpful not only for algorithm selection
and configuration, but further for selecting hard benchmark instances, that
distinguish different algorithms and to analyze hardness properties of problem
classes \cite[]{Hutter2014}.
\citeauthor{Hutter2014} propose \textit{empiricial performance models} (EPM) for runtime prediction.
These EPMs use a set of generic and problem-specific features to model the runtime characteristics.
Another study on runtime prediction for TSP has been published in \cite[]{Mersmann2013}, where the authors define a set of 47 TSP-specific features to asses instance hardness and algorithm performance.
For a comprehensive overview on literature in runtime prediction, we refer the interested reader to \cite[]{Hutter2014}.

The previously discussed work considered algorithm configuration and selection within one solver to optimize its performance. As mentioned earlier,
other approaches are focused towards combining multiple distinct solvers into a \textit{portfolio solver} \cite[]{Amadini2016a}.
Using machine learning and heuristics, the planning component of the portfolio solver determines the execution schedule of the solver \cite[]{Malitsky2012,Seipp2015}. 
In case of parallel portfolio solvers, a subset of solvers is run in parallel until a solution is found or, if the optimal solution is wanted, can exchange information about intermediate solutions found during search, e.g. sharing the best found objective bound \cite[]{Amadini2014c}.
Popular portfolio solvers include SATZilla \cite[]{Xu2007}, CPHydra \cite[]{OMahony2008}, Sunny-CP \cite[]{Amadini2014b}, or HaifaCSP \cite[]{Veksler2016}.

\subsection{Constraint Learning}

During the last decade, considerable progress has been made in the field of automatic constraint learning. Starting from a dataset of solutions and non-solutions examples, several approaches have been proposed to extract constraint models fitting the data. Pioneering this question, the ICON European project\footnote{\url{http://www.icon-fet.eu}} explored different approaches to this problem. It is worth noticing that these approaches to learn in CP are different from the previously described usages of predictive ML to CP, as, unlike statistical ML, the learning model is based on logic-driven approaches which extracts an exact model from examples.
Nevertheless, constraint learning approaches can be used to define predictive models to support other constraint models too, and are therefore included here as well.

In \cite[]{Beldiceanu2011,Beldiceanu2012}, Beldiceanu and Simonis have proposed {\sc ModelSeeker}, an approach that returns the best candidate global constraint representing a pattern occurring in a set of positive examples.
Following initial research ideas published in \cite[]{Bessiere2007}, \citeauthor{Bessiere2007} subsequently developed {\it Constraint Acquisition} as a strong inductive constraint learning framework \cite[]{Bessiere2015,Tsouros2019,Tsouros2020}. Starting from sequences of integers representing solution and non-solutions, constraint acquisition progressively refines a admissible and maximal model which accommodates all positive and negative examples. 
In \cite[]{Lallouet2007,Lallouet2010}, \citeauthor{Lallouet2007} had already proposed a constraint acquisition method based on inductive logic programming where both positive and negative examples can be handled, but the method captured the constraint network structure using some input background knowledge. 
Constraint Acquisition is independent of any background knowledge and just requires a bias, namely a subset of a constraint language, to be given as input. 
Interestingly, these constraint learning approaches are all derived from initial ideas developed in Inductive Logic Programming (ILP) \cite[]{DeRaedt2016}. 
The framework developed in this paper does not originate from ILP and does not try to infer a full CSP or Constraint Optimization model from sequences of positive and negative examples. 
Instead, it learns from existing solved instances to acquire suggested boundaries for the optimization variables. 
In that respect, it can complement Constraint Acquisition methods by exploiting solved instances and not only solutions and non-solutions.  

\subsection{Learning to Solve}
Applications and research on predictive ML for CP are sometimes classified as \textit{learning to solve}, putting an emphasis on the ML component and its contribution to CP.
These terminology is especially present in research that focuses on learning to solve optimization problems without the need for an additional solver \cite[]{Vinyals2015,Bello2017,Dai2017,Kumar2020,Cappart2021}.
Connected to the development of deep learning techniques, these approaches are able to solve small instances of constraint problems, but are not competitive to the capabilities of state-of-the-art constraint solvers.
A recent survey on the usage of reinforcement learning for combinatorial optimization can be found in \cite{Mazyavkina2021}.

\section{Estimating Objective Boundaries} \label{sec:boundaryestimation}

In this part of the paper, we present one application of predictive machine learning for constraint optimization, namely \method, a novel boundary estimation technique.
Boundary estimation supports the constraint solver by adding additional boundaries on the objective of a problem instance.
The objective boundaries are estimated via a machine learning model, that has been trained on previously solved problem instances.
Through the additional constraints, the search space of the constraint problem is pruned, which again allows to find good solutions early during search.

In general, exact solvers already include heuristics to find feasible initial solutions that can be
used for bounding the search \cite[]{Hooker2012}. For example, some CP and MIP
solvers use LP relaxations of the problem to find boundary. Other CP solvers rely on
good branching heuristics and constraint propagation to find close bounds
early \cite[]{Rossi2006}. These approaches are central to the modus operandi of
the solvers and crucial for their performance. 
Boundary estimation via predictive ML runs an additional bounding step before executing the solver and uses a ML-based heuristic, that is learned from historical data to already bound the objective and search space of the COP instance.

With boundary estimation, a different approach to COP solving is taken.
The CP solver exploits the constraint structure of a COP and considers only the current instance. 
In contrast, we train the ML model, which we refer to as the
\textit{estimator}, on the structure of instance parameters and the actual
objective value from example instances. Thus it only indirectly infers the model
constraints, but it is not explicitly made aware of them. Our approach combines
data- and logic-driven approaches to solve COPs and benefits from the
estimation provided by the data-driven prediction and also the optimal solution
computed by the logic-driven COP solver. In principle, \method boosts
the solving process by reducing the search space with estimated tight boundaries
on the optimal objective.

We now introduce the concept of estimated boundaries, 
which refers to providing close lower and upper bounds
for the optimal value \zopt of \fz.

\begin{Definition}[Estimation]
An \emph{estimation} is a domain $\estzlb..\estzub$ which defines boundaries
for the domain of $\fz$.
The domain boundaries are predicted by a supervised ML model $P : \mathbb{R}^n \rightarrow \mathbb{R}^2$, that is, $\langle \estzlb, \estzub \rangle = P(\mathbf{x})$.
\end{Definition}

\begin{Definition}[Admissible/Inadmissible Estimations]
    An estimation $\estzlb..\estzub$ is \emph{admissible} iff $\zopt \in \estzlb..\estzub$. %
    Otherwise, the estimation is said to be inadmissible.
\end{Definition}

We further classify the two domain boundaries as \emph{cutting} and
\emph{limiting} boundaries in relation to their effect on the solver's search process.
Depending on whether the COP is a minimization or maximization problem, these
terms refer to different domain boundaries.

\begin{Definition}[Cutting Boundary]
The \emph{cutting boundary} is the domain boundary that reduces the number of reachable solutions. 
For minimization, this is the upper domain boundary \ub{\z}; 
for maximization, the lower domain boundary \lb{\z}.
\end{Definition}

\begin{Definition}[Limiting Boundary]
The \emph{limiting boundary} is the domain boundary that does not reduce the number of reachable solutions, but only reduces the search space to be explored.
For minimization, this is the lower domain boundary \lb{\z}; 
for maximization, the upper domain boundary \ub{\z}.
\end{Definition}

For the sake of simplicity, in the rest of the paper, we focus exclusively on minimization
problems, however, \method is similarly applicable to maximization
problems.

\subsection{Optimization with Boundary Constraints}
\label{sec:process}

We first present the full process to solve a COP with Bion, which receives as inputs both an
optimization model, describing the problem in terms of the variables \vars and
constraints \constraints, and its instance parameters which include data
structure sizes, boundaries and constraints parameters. The COP is the same for
all instances, only the parameters given as a separate input can change.
The process of solving COPs with an already trained estimator is shown as a pseudocode formulation in Algorithm \ref{alg:bion}.
For simplicity of the formulation, we represent the static COP and the instance parameters merged into one triple $\langle \vars, \constraints, f_\obj \rangle$.

\begin{algorithm}[t]
\begin{algorithmic}[1]
\Function{SolveWithBion}{COP Instance $\langle \vars, \constraints, f_\obj \rangle$, Estimator $P$, Solver $S$}
\State $x \gets$ \Call{preprocess}{$\langle \vars, \constraints, f_\obj \rangle$} \Comment{Extract feature vector $x$ from COP instance}
\State $\langle \estzlb, \estzub \rangle \gets P(x)$ \Comment{Predict objective boundary} 
\State $\constraints'  \gets \constraints \cup \{z \in [\estzlb, \estzub]\}$ \Comment{Update COP with boundary constraint} 
\State Result $\gets$ \Call{solve}{$\langle \vars, \constraints', f_\obj \rangle$} \Comment{Solve updated COP with CP solver} 

\If{Result = Unsatisfiable}
    \State $\constraints'' \gets \constraints \cup \{z \notin [\estzlb, \estzub]\}$ \Comment{Update COP with negated boundary constraint} 
    \State Result $\gets$ \Call{solve}{$\langle \vars, \constraints'', f_\obj \rangle$} \Comment{Solve updated COP with CP solver} 
\EndIf
\State \Return Result, $x, z$ \Comment{Return solver result; include $x$ and $z$ for future model training}
\EndFunction
\end{algorithmic}
\caption{Pseudocode formulation for the COP solving process with Bion}
\label{alg:bion}
\end{algorithm}

Boundary estimation adds a preprocessing step to COP solving, as well as a rule
for handling unsatisfiable instances. During preprocessing, the current problem
instance is analyzed and the trained estimator estimates a boundary on the
objective value of this specific instance. To provide the estimated boundary,
instance-specific features are extracted from the COP model and its instance
parameters. These features serve as the input of the estimator, which returns
the estimated boundary value.

Afterwards, \method adds the boundary value as an additional constraint on the
objective variable to the optimization model. The extended, now
instance-specific model and the unmodified instance parameters are then given to
the solver. If the solver returns a solution, the process ends, as the problem
is solved. However, if the approximated objective value is too low, i.e. the
estimation is inadmissible, it can render the problem unsatisfiable. In this
case, \method restarts the solver with the inverted boundary constraint, such
that the estimation is a lower bound on \obj. If the COP is now satisfiable, the
estimation was inadmissible. Otherwise, unsatisfiability is due to other
reasons. When the COP is solved, both the input and the objective value
are stored for future training of the estimator.

\subsection{Feature Selection} \label{sec:features}

Each COP consists of constraints, variables, and their domains. To use these
components as estimator inputs, it is necessary to extract and transform
features, which describe a problem and its data in a meaningful way. As we discussed, 
a good set of features captures relations between instance data and the quantity to
estimate, i.e., the objective value. Furthermore, the features are dynamic in
terms of the individual variability of an instance, but static in the number of
features, that is, each instance of one problem is represented by the same
features.

Among the various types of variable structures one can find in COPs, non-fixed
data structures such as arrays, lists or sets, are the main contributors of
variability and of major relevance for feature selection. For each of these
data structures, $9$ common metrics from the main categories of descriptive
statistics are calculated to gather an abstract, but comprehensive description
of the contained data: (1) the number of elements, including the (2) minimum and
(3) maximum values. The central tendency of the data is described by both (4)
arithmetic mean and (5) median, the dispersion by the (6) standard deviation and
(7) interquartile range. Finally, (8) skewness and (9) kurtosis describe the
shape of the data distribution. Scalar variables are each added as
features with their value.

The constraints of a model are fixed for all instances, but due to different
data inputs, the inferred final models for each problem instance differ. To
capture information about the variables generated during compilation and
model-specific, we use $95$ features as implemented in the current version of
\texttt{mzn2feat} \cite[]{Amadini2014}, which are listed in Table \ref{tab:instances}. 
Several of these features have the same value for all instances as they capture static properties
of the COP, but some add useful information for different instances, which
supports the learning performance.
Nevertheless, preliminary experiments showed that including the non-static features of the constraint model improves estimation accuracy.

In general, all features can have varying relevance to express the data,
depending on the model and the necessary input. However, as \method is designed
to be problem-independent, the same features are at first calculated for all
problems. During the data preprocessing phase of the model training, all
features with zero variance, that is, the same value for all instances, are
removed to reduce the model complexity. Each feature is further standardized by
subtracting the mean and dividing by the standard deviation.

\subsection{Avoiding Inadmissible Estimations} \label{sec:training}

One main risk, when automatically adding constraints to a COP, is to render the
problem unsatisfiable. In the case of \BE, an inadmissible estimation below the
optimum value prohibits the solver from finding any feasible solution. Even
though this is often detected early by the solver, there is no guarantee and
thus it is necessary to tame this issue. To mitigate the risk, three
counter-measures are considered in the design of \method.

First, during COP solving with \method, unsatisfiable instances provide information about the problem and allow
us to restart the solver with an inverted boundary constraint, such that an upper
bound becomes a lower bound; Second, training the estimator with an asymmetric
loss function penalizes errors on the side of inadmissible underestimates
stronger than admissible overestimates, and thereby discourages misestimations;
Third, besides exploiting the training error, the estimator is explicitly
trained to overestimate. This requires adjusting the training label from the
actual optimal objective value towards an overestimation.

\subsubsection{Symmetric vs. Asymmetric Loss} \label{sec:losses}

\begin{figure}[t]
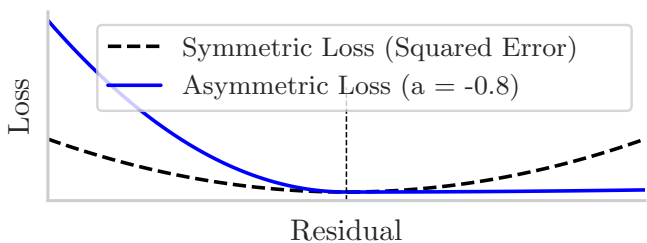

  \centering
  \resizebox{0.5\textwidth}{!}{%
\begingroup%
\makeatletter%
\begin{pgfpicture}%
\pgfpathrectangle{\pgfpointorigin}{\pgfqpoint{2.979679in}{1.103342in}}%
\pgfusepath{use as bounding box, clip}%
\begin{pgfscope}%
\pgfsetbuttcap%
\pgfsetmiterjoin%
\definecolor{currentfill}{rgb}{1.000000,1.000000,1.000000}%
\pgfsetfillcolor{currentfill}%
\pgfsetlinewidth{0.000000pt}%
\definecolor{currentstroke}{rgb}{1.000000,1.000000,1.000000}%
\pgfsetstrokecolor{currentstroke}%
\pgfsetdash{}{0pt}%
\pgfpathmoveto{\pgfqpoint{0.000000in}{0.000000in}}%
\pgfpathlineto{\pgfqpoint{2.979679in}{0.000000in}}%
\pgfpathlineto{\pgfqpoint{2.979679in}{1.103342in}}%
\pgfpathlineto{\pgfqpoint{0.000000in}{1.103342in}}%
\pgfpathclose%
\pgfusepath{fill}%
\end{pgfscope}%
\begin{pgfscope}%
\pgfsetbuttcap%
\pgfsetmiterjoin%
\definecolor{currentfill}{rgb}{1.000000,1.000000,1.000000}%
\pgfsetfillcolor{currentfill}%
\pgfsetlinewidth{0.000000pt}%
\definecolor{currentstroke}{rgb}{0.000000,0.000000,0.000000}%
\pgfsetstrokecolor{currentstroke}%
\pgfsetstrokeopacity{0.000000}%
\pgfsetdash{}{0pt}%
\pgfpathmoveto{\pgfqpoint{0.216536in}{0.216536in}}%
\pgfpathlineto{\pgfqpoint{2.970079in}{0.216536in}}%
\pgfpathlineto{\pgfqpoint{2.970079in}{1.093742in}}%
\pgfpathlineto{\pgfqpoint{0.216536in}{1.093742in}}%
\pgfpathclose%
\pgfusepath{fill}%
\end{pgfscope}%
\begin{pgfscope}%
\definecolor{textcolor}{rgb}{0.150000,0.150000,0.150000}%
\pgfsetstrokecolor{textcolor}%
\pgfsetfillcolor{textcolor}%
\pgftext[x=1.593308in,y=0.129036in,,top]{\color{textcolor}\rmfamily\fontsize{9.600000}{11.520000}\selectfont Residual}%
\end{pgfscope}%
\begin{pgfscope}%
\definecolor{textcolor}{rgb}{0.150000,0.150000,0.150000}%
\pgfsetstrokecolor{textcolor}%
\pgfsetfillcolor{textcolor}%
\pgftext[x=0.129036in,y=0.655139in,,bottom,rotate=90.000000]{\color{textcolor}\rmfamily\fontsize{9.600000}{11.520000}\selectfont Loss}%
\end{pgfscope}%
\begin{pgfscope}%
\pgfpathrectangle{\pgfqpoint{0.216536in}{0.216536in}}{\pgfqpoint{2.753543in}{0.877205in}}%
\pgfusepath{clip}%
\pgfsetbuttcap%
\pgfsetroundjoin%
\pgfsetlinewidth{1.204500pt}%
\definecolor{currentstroke}{rgb}{0.000000,0.000000,0.000000}%
\pgfsetstrokecolor{currentstroke}%
\pgfsetdash{{4.440000pt}{1.920000pt}}{0.000000pt}%
\pgfpathmoveto{\pgfqpoint{0.216536in}{0.502539in}}%
\pgfpathlineto{\pgfqpoint{0.262582in}{0.486351in}}%
\pgfpathlineto{\pgfqpoint{0.308628in}{0.470713in}}%
\pgfpathlineto{\pgfqpoint{0.354674in}{0.455626in}}%
\pgfpathlineto{\pgfqpoint{0.400720in}{0.441090in}}%
\pgfpathlineto{\pgfqpoint{0.446766in}{0.427104in}}%
\pgfpathlineto{\pgfqpoint{0.492812in}{0.413669in}}%
\pgfpathlineto{\pgfqpoint{0.538857in}{0.400784in}}%
\pgfpathlineto{\pgfqpoint{0.584903in}{0.388450in}}%
\pgfpathlineto{\pgfqpoint{0.630949in}{0.376667in}}%
\pgfpathlineto{\pgfqpoint{0.676995in}{0.365435in}}%
\pgfpathlineto{\pgfqpoint{0.723041in}{0.354753in}}%
\pgfpathlineto{\pgfqpoint{0.769087in}{0.344621in}}%
\pgfpathlineto{\pgfqpoint{0.815133in}{0.335040in}}%
\pgfpathlineto{\pgfqpoint{0.861179in}{0.326010in}}%
\pgfpathlineto{\pgfqpoint{0.907224in}{0.317531in}}%
\pgfpathlineto{\pgfqpoint{0.953270in}{0.309602in}}%
\pgfpathlineto{\pgfqpoint{0.999316in}{0.302224in}}%
\pgfpathlineto{\pgfqpoint{1.045362in}{0.295396in}}%
\pgfpathlineto{\pgfqpoint{1.091408in}{0.289119in}}%
\pgfpathlineto{\pgfqpoint{1.137454in}{0.283392in}}%
\pgfpathlineto{\pgfqpoint{1.183500in}{0.278217in}}%
\pgfpathlineto{\pgfqpoint{1.229546in}{0.273591in}}%
\pgfpathlineto{\pgfqpoint{1.275591in}{0.269517in}}%
\pgfpathlineto{\pgfqpoint{1.321637in}{0.265993in}}%
\pgfpathlineto{\pgfqpoint{1.367683in}{0.263019in}}%
\pgfpathlineto{\pgfqpoint{1.413729in}{0.260597in}}%
\pgfpathlineto{\pgfqpoint{1.459775in}{0.258725in}}%
\pgfpathlineto{\pgfqpoint{1.505821in}{0.257403in}}%
\pgfpathlineto{\pgfqpoint{1.551867in}{0.256632in}}%
\pgfpathlineto{\pgfqpoint{1.597912in}{0.256412in}}%
\pgfpathlineto{\pgfqpoint{1.643958in}{0.256742in}}%
\pgfpathlineto{\pgfqpoint{1.690004in}{0.257623in}}%
\pgfpathlineto{\pgfqpoint{1.736050in}{0.259055in}}%
\pgfpathlineto{\pgfqpoint{1.782096in}{0.261037in}}%
\pgfpathlineto{\pgfqpoint{1.828142in}{0.263570in}}%
\pgfpathlineto{\pgfqpoint{1.874188in}{0.266654in}}%
\pgfpathlineto{\pgfqpoint{1.920234in}{0.270288in}}%
\pgfpathlineto{\pgfqpoint{1.966279in}{0.274472in}}%
\pgfpathlineto{\pgfqpoint{2.012325in}{0.279208in}}%
\pgfpathlineto{\pgfqpoint{2.058371in}{0.284494in}}%
\pgfpathlineto{\pgfqpoint{2.104417in}{0.290330in}}%
\pgfpathlineto{\pgfqpoint{2.150463in}{0.296717in}}%
\pgfpathlineto{\pgfqpoint{2.196509in}{0.303655in}}%
\pgfpathlineto{\pgfqpoint{2.242555in}{0.311144in}}%
\pgfpathlineto{\pgfqpoint{2.288601in}{0.319183in}}%
\pgfpathlineto{\pgfqpoint{2.334646in}{0.327772in}}%
\pgfpathlineto{\pgfqpoint{2.380692in}{0.336913in}}%
\pgfpathlineto{\pgfqpoint{2.426738in}{0.346603in}}%
\pgfpathlineto{\pgfqpoint{2.472784in}{0.356845in}}%
\pgfpathlineto{\pgfqpoint{2.518830in}{0.367637in}}%
\pgfpathlineto{\pgfqpoint{2.564876in}{0.378980in}}%
\pgfpathlineto{\pgfqpoint{2.610922in}{0.390873in}}%
\pgfpathlineto{\pgfqpoint{2.656968in}{0.403317in}}%
\pgfpathlineto{\pgfqpoint{2.703013in}{0.416312in}}%
\pgfpathlineto{\pgfqpoint{2.749059in}{0.429857in}}%
\pgfpathlineto{\pgfqpoint{2.795105in}{0.443953in}}%
\pgfpathlineto{\pgfqpoint{2.841151in}{0.458599in}}%
\pgfpathlineto{\pgfqpoint{2.887197in}{0.473796in}}%
\pgfpathlineto{\pgfqpoint{2.933243in}{0.489544in}}%
\pgfpathlineto{\pgfqpoint{2.970079in}{0.502539in}}%
\pgfpathlineto{\pgfqpoint{2.970079in}{0.502539in}}%
\pgfusepath{stroke}%
\end{pgfscope}%
\begin{pgfscope}%
\pgfpathrectangle{\pgfqpoint{0.216536in}{0.216536in}}{\pgfqpoint{2.753543in}{0.877205in}}%
\pgfusepath{clip}%
\pgfsetroundcap%
\pgfsetroundjoin%
\pgfsetlinewidth{1.204500pt}%
\definecolor{currentstroke}{rgb}{0.000000,0.000000,1.000000}%
\pgfsetstrokecolor{currentstroke}%
\pgfsetdash{}{0pt}%
\pgfpathmoveto{\pgfqpoint{0.216536in}{1.053869in}}%
\pgfpathlineto{\pgfqpoint{0.244164in}{1.022185in}}%
\pgfpathlineto{\pgfqpoint{0.271791in}{0.991143in}}%
\pgfpathlineto{\pgfqpoint{0.299419in}{0.960744in}}%
\pgfpathlineto{\pgfqpoint{0.327046in}{0.930987in}}%
\pgfpathlineto{\pgfqpoint{0.354674in}{0.901872in}}%
\pgfpathlineto{\pgfqpoint{0.382302in}{0.873399in}}%
\pgfpathlineto{\pgfqpoint{0.409929in}{0.845568in}}%
\pgfpathlineto{\pgfqpoint{0.437557in}{0.818380in}}%
\pgfpathlineto{\pgfqpoint{0.465184in}{0.791834in}}%
\pgfpathlineto{\pgfqpoint{0.492812in}{0.765930in}}%
\pgfpathlineto{\pgfqpoint{0.520439in}{0.740669in}}%
\pgfpathlineto{\pgfqpoint{0.548067in}{0.716050in}}%
\pgfpathlineto{\pgfqpoint{0.575694in}{0.692072in}}%
\pgfpathlineto{\pgfqpoint{0.603322in}{0.668738in}}%
\pgfpathlineto{\pgfqpoint{0.630949in}{0.646045in}}%
\pgfpathlineto{\pgfqpoint{0.658577in}{0.623995in}}%
\pgfpathlineto{\pgfqpoint{0.686204in}{0.602587in}}%
\pgfpathlineto{\pgfqpoint{0.713832in}{0.581821in}}%
\pgfpathlineto{\pgfqpoint{0.741459in}{0.561697in}}%
\pgfpathlineto{\pgfqpoint{0.769087in}{0.542216in}}%
\pgfpathlineto{\pgfqpoint{0.796714in}{0.523377in}}%
\pgfpathlineto{\pgfqpoint{0.824342in}{0.505180in}}%
\pgfpathlineto{\pgfqpoint{0.851969in}{0.487625in}}%
\pgfpathlineto{\pgfqpoint{0.879597in}{0.470713in}}%
\pgfpathlineto{\pgfqpoint{0.907224in}{0.454443in}}%
\pgfpathlineto{\pgfqpoint{0.934852in}{0.438815in}}%
\pgfpathlineto{\pgfqpoint{0.962479in}{0.423829in}}%
\pgfpathlineto{\pgfqpoint{0.990107in}{0.409486in}}%
\pgfpathlineto{\pgfqpoint{1.017735in}{0.395785in}}%
\pgfpathlineto{\pgfqpoint{1.045362in}{0.382726in}}%
\pgfpathlineto{\pgfqpoint{1.072990in}{0.370309in}}%
\pgfpathlineto{\pgfqpoint{1.100617in}{0.358535in}}%
\pgfpathlineto{\pgfqpoint{1.128245in}{0.347402in}}%
\pgfpathlineto{\pgfqpoint{1.155872in}{0.336913in}}%
\pgfpathlineto{\pgfqpoint{1.183500in}{0.327065in}}%
\pgfpathlineto{\pgfqpoint{1.211127in}{0.317859in}}%
\pgfpathlineto{\pgfqpoint{1.238755in}{0.309296in}}%
\pgfpathlineto{\pgfqpoint{1.266382in}{0.301375in}}%
\pgfpathlineto{\pgfqpoint{1.294010in}{0.294096in}}%
\pgfpathlineto{\pgfqpoint{1.321637in}{0.287460in}}%
\pgfpathlineto{\pgfqpoint{1.349265in}{0.281466in}}%
\pgfpathlineto{\pgfqpoint{1.376892in}{0.276114in}}%
\pgfpathlineto{\pgfqpoint{1.404520in}{0.271404in}}%
\pgfpathlineto{\pgfqpoint{1.432147in}{0.267336in}}%
\pgfpathlineto{\pgfqpoint{1.459775in}{0.263911in}}%
\pgfpathlineto{\pgfqpoint{1.487402in}{0.261128in}}%
\pgfpathlineto{\pgfqpoint{1.515030in}{0.258987in}}%
\pgfpathlineto{\pgfqpoint{1.542657in}{0.257489in}}%
\pgfpathlineto{\pgfqpoint{1.570285in}{0.256632in}}%
\pgfpathlineto{\pgfqpoint{1.597912in}{0.256409in}}%
\pgfpathlineto{\pgfqpoint{1.800514in}{0.256632in}}%
\pgfpathlineto{\pgfqpoint{2.003116in}{0.257282in}}%
\pgfpathlineto{\pgfqpoint{2.205718in}{0.258357in}}%
\pgfpathlineto{\pgfqpoint{2.408320in}{0.259859in}}%
\pgfpathlineto{\pgfqpoint{2.610922in}{0.261788in}}%
\pgfpathlineto{\pgfqpoint{2.813523in}{0.264143in}}%
\pgfpathlineto{\pgfqpoint{2.970079in}{0.266254in}}%
\pgfpathlineto{\pgfqpoint{2.970079in}{0.266254in}}%
\pgfusepath{stroke}%
\end{pgfscope}%
\begin{pgfscope}%
\pgfpathrectangle{\pgfqpoint{0.216536in}{0.216536in}}{\pgfqpoint{2.753543in}{0.877205in}}%
\pgfusepath{clip}%
\pgfsetbuttcap%
\pgfsetroundjoin%
\pgfsetlinewidth{0.501875pt}%
\definecolor{currentstroke}{rgb}{0.000000,0.000000,0.000000}%
\pgfsetstrokecolor{currentstroke}%
\pgfsetdash{{1.850000pt}{0.800000pt}}{0.000000pt}%
\pgfpathmoveto{\pgfqpoint{1.593308in}{0.216536in}}%
\pgfpathlineto{\pgfqpoint{1.593308in}{0.742860in}}%
\pgfusepath{stroke}%
\end{pgfscope}%
\begin{pgfscope}%
\pgfsetrectcap%
\pgfsetmiterjoin%
\pgfsetlinewidth{1.003750pt}%
\definecolor{currentstroke}{rgb}{0.800000,0.800000,0.800000}%
\pgfsetstrokecolor{currentstroke}%
\pgfsetdash{}{0pt}%
\pgfpathmoveto{\pgfqpoint{0.216536in}{0.216536in}}%
\pgfpathlineto{\pgfqpoint{0.216536in}{1.093742in}}%
\pgfusepath{stroke}%
\end{pgfscope}%
\begin{pgfscope}%
\pgfsetrectcap%
\pgfsetmiterjoin%
\pgfsetlinewidth{1.003750pt}%
\definecolor{currentstroke}{rgb}{0.800000,0.800000,0.800000}%
\pgfsetstrokecolor{currentstroke}%
\pgfsetdash{}{0pt}%
\pgfpathmoveto{\pgfqpoint{0.216536in}{0.216536in}}%
\pgfpathlineto{\pgfqpoint{2.970079in}{0.216536in}}%
\pgfusepath{stroke}%
\end{pgfscope}%
\begin{pgfscope}%
\pgfsetbuttcap%
\pgfsetmiterjoin%
\definecolor{currentfill}{rgb}{1.000000,1.000000,1.000000}%
\pgfsetfillcolor{currentfill}%
\pgfsetfillopacity{0.800000}%
\pgfsetlinewidth{0.803000pt}%
\definecolor{currentstroke}{rgb}{0.800000,0.800000,0.800000}%
\pgfsetstrokecolor{currentstroke}%
\pgfsetstrokeopacity{0.800000}%
\pgfsetdash{}{0pt}%
\pgfpathmoveto{\pgfqpoint{0.468688in}{0.633714in}}%
\pgfpathlineto{\pgfqpoint{2.884524in}{0.633714in}}%
\pgfpathquadraticcurveto{\pgfqpoint{2.908968in}{0.633714in}}{\pgfqpoint{2.908968in}{0.658159in}}%
\pgfpathlineto{\pgfqpoint{2.908968in}{1.008186in}}%
\pgfpathquadraticcurveto{\pgfqpoint{2.908968in}{1.032631in}}{\pgfqpoint{2.884524in}{1.032631in}}%
\pgfpathlineto{\pgfqpoint{0.468688in}{1.032631in}}%
\pgfpathquadraticcurveto{\pgfqpoint{0.444244in}{1.032631in}}{\pgfqpoint{0.444244in}{1.008186in}}%
\pgfpathlineto{\pgfqpoint{0.444244in}{0.658159in}}%
\pgfpathquadraticcurveto{\pgfqpoint{0.444244in}{0.633714in}}{\pgfqpoint{0.468688in}{0.633714in}}%
\pgfpathclose%
\pgfusepath{stroke,fill}%
\end{pgfscope}%
\begin{pgfscope}%
\pgfsetbuttcap%
\pgfsetroundjoin%
\pgfsetlinewidth{1.204500pt}%
\definecolor{currentstroke}{rgb}{0.000000,0.000000,0.000000}%
\pgfsetstrokecolor{currentstroke}%
\pgfsetdash{{4.440000pt}{1.920000pt}}{0.000000pt}%
\pgfpathmoveto{\pgfqpoint{0.493133in}{0.933659in}}%
\pgfpathlineto{\pgfqpoint{0.737577in}{0.933659in}}%
\pgfusepath{stroke}%
\end{pgfscope}%
\begin{pgfscope}%
\definecolor{textcolor}{rgb}{0.150000,0.150000,0.150000}%
\pgfsetstrokecolor{textcolor}%
\pgfsetfillcolor{textcolor}%
\pgftext[x=0.835355in,y=0.890882in,left,base]{\color{textcolor}\rmfamily\fontsize{8.800000}{10.560000}\selectfont Symmetric Loss (Squared Error)}%
\end{pgfscope}%
\begin{pgfscope}%
\pgfsetroundcap%
\pgfsetroundjoin%
\pgfsetlinewidth{1.204500pt}%
\definecolor{currentstroke}{rgb}{0.000000,0.000000,1.000000}%
\pgfsetstrokecolor{currentstroke}%
\pgfsetdash{}{0pt}%
\pgfpathmoveto{\pgfqpoint{0.493133in}{0.752534in}}%
\pgfpathlineto{\pgfqpoint{0.737577in}{0.752534in}}%
\pgfusepath{stroke}%
\end{pgfscope}%
\begin{pgfscope}%
\definecolor{textcolor}{rgb}{0.150000,0.150000,0.150000}%
\pgfsetstrokecolor{textcolor}%
\pgfsetfillcolor{textcolor}%
\pgftext[x=0.835355in,y=0.709757in,left,base]{\color{textcolor}\rmfamily\fontsize{8.800000}{10.560000}\selectfont Asymmetric Loss (a = -0.8)}%
\end{pgfscope}%
\end{pgfpicture}%
\makeatother%
\endgroup%
}
  \caption{Quadratic symmetric and asymmetric loss functions. The asymmetric loss assigns
    a higher loss to a negative residuals, but lower loss to overestimations,
    than the symmetric squared error loss.}
  \label{fig:loss_functions}
\end{figure}

Common loss functions used for training ML models, such as MSE, are symmetric and do not differentiate between positive or negative
errors. An asymmetric loss function, on the other hand, assigns higher loss
values for either under- or overestimations, which means that certain errors are
more penalized than others. Figure~\ref{fig:loss_functions} shows an example of
quadratic symmetric and asymmetric loss functions.

\textit{Shifted Squared Error Loss} is an imbalanced variant of squared error
loss. The parameter \(a\) shifts the penalization towards under- or
overestimation and influences the magnitude of the penalty. Formally speaking,
\begin{Definition}[Shifted Squared Error Loss]
\[ L(r) = r^2 \cdot (sgn(r) + a)^2 \text{ with absolute error } r = \hat{y} - y\]
where $\hat{y}$ is the estimated value and $y$ is the true target value, 
and $a$ is a parameter which shifts the penalization towards under- or overestimation.
\end{Definition}

In Section~\ref{sec:experiments}, we compare the usage of both symmetric and
asymmetric loss functions plus label shift to evaluate the importance of
adjusting model training to the problem instances.

\subsubsection{Label Shift} \label{sec:labelshift}

Boundary estimation only approximates the objective function of the COP, which
means there is no requirement on the convergence towards a truly optimum
solution. 
Said otherwise, \method can accept estimation errors after training. This allows
\method to work with fewer training examples than what could be expected with a
desired exact training method. This is appropriate here, as collecting labeled
data requires solving many COPs instances first, which can be very costly.

However, there is a risk that the trained model underestimates (in case of
minimization) the actual objective value. Such an inadmissible estimation, as
defined above, leads to an unsatisfiable constraint system and prohibits the COP
solver from finding any feasible solution. On the other hand, a too loose, but
admissible overestimation may not sufficiently approximate the actual optimum.
To address this risk, we introduce {\it label shift} in \method, which adjusts the
training procedure with one user-controlled parameter. Label shift is similar to
the concept of prediction shift described in \cite[]{Tolstikov2017}, but based on
the specific COP model.

\begin{Definition}[Label Shift]
\begin{align*}
y' &= y + \lambda \,(\ub{z} - y)\,\quad \text{\emph{(Overestimation)}}\\
y' &= y - \lambda \,(y - \lb{z})\,\quad \text{\emph{(Underestimation)}}
\end{align*}
\end{Definition}
where \(\ub{z}\) is the upper bound of the objective domain,
\(y\) is the optimal objective value of the training instance, and
\(y'\) is the adjusted label for training the estimator, as the result of the label shift adjustment. 
Label shift depends on \(\lambda\), which is an adjustment factor to shift the target value
\(y\) between the domain boundary and the actual optimum. The trade-off between
a close and admissible estimation and an inadmissible estimation is thus
controlled by the value of $\lambda$.

\subsection{Estimated Boundaries during Search}  %
\label{sec:method_cp}

Using \method to solve a COP consists of the following steps:

\begin{enumerate}
    \item (Initially) Train an estimator model for the COP
    \item Extract a feature vector from each COP instance
    \item Estimate both a lower and an upper objective boundaries
    \item Update the COP with estimated boundaries
    \item Solve the updated COP with the solver
\end{enumerate}

The boundaries provided by the estimator can be embedded as hard constraints on the objective variable, i.e., by adding $\z \in \estzlb \dots \estzub$.
The induced overhead is negligible, but dealing with misestimations requires additional control.
If all feasible solutions are excluded, because the cutting bound is wrongly estimated, the instance is rendered unsatisfiable.
This issue is handled by reverting to the original domain.

If only optimal solutions are excluded, because the limiting bound is wrongly estimated, then only non-optimal solutions can be returned and this stays impossible to notice.
This issue cannot be detected in a single optimization run of the solver. 
However, in practical cases where the goal is to find good-enough solutions early rather than finding truly-proven optima, it can be an acceptable risk to come-up with an good approximation of the optimal solutions only.

Conclusively, hard boundary constraints are especially suited for cases where a high confidence in the quality of the estimator has been gained, and the occurrence of inadmissible estimations is unlikely.

\section{Experiments} \label{sec:experiments}

We experimentally evaluate our method \method in three experiments, which focus on the impact of label shift and asymmetric loss functions for training the estimator, on
the estimators' performance to bound the objective domain, 
and on the impact of estimated boundaries on solver performance.

\subsection{Setup} \label{sec:expsetup}
\subsubsection{Constraint Optimization Problems}

We selected seven problems, that were previously used
in MiniZinc challenges \cite{Stuckey2014} to evaluate CP solvers. These
problems are those with the largest number of instances in the MiniZinc
benchmarks\footnote{Accessible at
  https://github.com/MiniZinc/minizinc-benchmarks/}, ranging from 50 to multiple
thousands. 
In practice, it is more likely to only have few examples instances
available, therefore we also include problems with few training examples. 
These COPs, which are all minimization problems, are listed in Table~\ref{tab:instances} along with the type of objective function, whether they contain a custom search strategy, and the number of available instances.
Considering training sets of different sizes, from 50 to over 11,000 instances, is relevant to understand scenarios that can benefit from boundary estimation.
The column \emph{Objective} describes the objective function type, which is related to the solver's ability to efficiently propagate domain boundaries.
The COPs have two main groups of objective functions. 
One group minimizes the sum, the other minimizes the maximum of a list of variables.
For the models minimizing the maximum, two formulations are present in our evaluation models. 
\emph{Max-Max} uses the MiniZinc built-in \textrm{max} ($\z = max(V)$), whereas \emph{Leq-Max} constraints the objective to be greater-or-equal all variables ($\forall\,\mathbf{v} \in V\,:\,\mathbf{v} \leq \z$). 
Both formulations are decomposed into different FlatZinc constraints, which
can have an impact on the ability to propagate constraints.

\begin{table}[t]
  \centering
  \small
  \caption{Overview of benchmark problems. All considered problems are
    minimization problems with a large variety in the number of available
    training instances.\label{tab:instances}}
\begin{tabularx}{\columnwidth}{XXXr}
\toprule
       Problem &  Objective & Search & Instances \\
\midrule
        MRCPSP &  Max-Max & Model-Specific &   11182 \\ %
         RCPSP &  Leq-Max & Model-Specific &     2904 \\ %
   Bin Packing &  Sum & Solver-Default &      500 \\ %
 Cutting Stock &  Sum & Solver-Default &      121 \\ %
       Jobshop &  Leq-Max & Solver-Default &       74 \\ %
           VRP &  Sum & Model-Specific &       74 \\ %
   Open Stacks &  Max-Max & Model-Specific &       50 \\ %
\bottomrule
\end{tabularx}
 \end{table}

\subsubsection{Training Settings}
We implement \method in Python using scikit-learn \cite[]{scikit-learn}. 
Exceptions are NNs, where
Keras \cite[]{Chollet2015} and TensorFlow \cite[]{abadi2016tensorflow} are used, and GTB, where
XGBoost \cite[]{Chen2016} is used, both to support custom loss functions.
mzn2feat \cite[]{Amadini2014} is used for COP feature extraction.
In our comparison, we consider asymmetric (\gtba, \nna)
and symmetric versions (\gtbs, \nns) of GTB and NN, and symmetric versions of SVM and linear regression.

The performed experiments are targeted towards evaluating the general
effectiveness of \BE over a range of different problems. Therefore, we used the
default parameters of each ML model as provided by the chosen frameworks as much
as possible. As loss factors for the asymmetric loss functions, we set \(a =
-1\) for \gtba and \(a = -0.8\) for \nna, where a smaller \(a\) caused problems
during training. The model parameters were not adjusted for individual problems.
Parameter tuning is also often not performed in a practical application,
although it potentially allows to improve the performance for some problems. For
Bin Packing, we introduced a trivial upper boundary as there was originally
no finite boundary in the model.

\subsection{Boundary Estimation Performance} \label{sec:mlresults}
We evaluate the capability to learn close and admissible objective boundaries,
that prune the objective variable's domain.
To this end, we focus on 
1) the impact of label shift and asymmetric loss functions for training the estimator, 
2) the estimators' performance to bound the objective domain, and 
3) the impact of estimated boundaries on solver performance.

Estimation performance is measured through repeated 10-fold validation. In each
repetition, the instances are randomly split into ten folds. Nine of these folds
form the training set, the other the validation set. The model is trained 10
times, one time with each fold as validation set. We repeat this process 30
times, to account for stochastic effects, and report median values.

Training times for the models are dependent on the number of training samples
available. We trained on commodity hardware without GPU acceleration.
In general, training takes less than five seconds per model, except for MRCPSP
with up to 30 seconds. Another exception are the neural networks which take on average 1
minute to train and maximum 6 minutes for MRCPSP.

\subsubsection{Adjustment Factors for Label Shift}
To avoid inadmissible estimations, we introduced the label shift method. Label
shift trains the estimator not on the exact objective value, but adjusts the
label of the training samples according to a configuration parameter
\(\lambda\).

\begin{figure}[H]
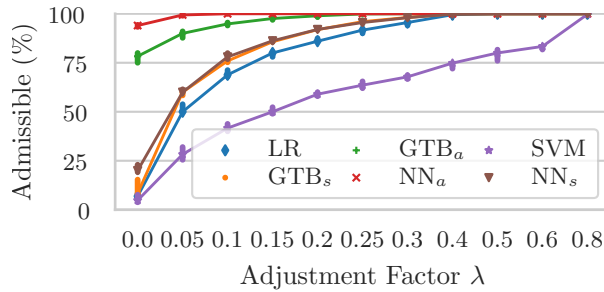

  \centering
\begingroup%
\makeatletter%
%
\makeatother%
\endgroup%
   \caption{Admissible estimations in relation to adjustment factor $\lambda$. Label shift increases admissible estimations for both symmetric and asymmetric models.
  }
  \label{fig:labelshift}
\end{figure}

We evaluate the effect of the adjustment factor (\(\lambda \in \{0, 0.1, 0.2,
\dots, 1.0\}\)) on the quality of estimations. The results are shown in
Figure~\ref{fig:labelshift}, both the ratio of admissible estimations and the
ratio of instances for which the domain is pruned. Here, we do not distinguish
the different benchmark problems, but aggregate the results as they show similar
behavior for the different problems.

The results confirm that choosing \(\lambda\) is a trade-off, where a larger
value increases admissible estimations, but reduces pruning. Furthermore, the
difference between symmetric and asymmetric loss is visible. Without label
shift, the symmetric models have close to 50\% admissible estimations, which is
expected as the symmetric error is equally distributed between inadmissible
underestimations and admissible overestimations. The asymmetric models achieve
\(88/92\%\) (\gtba/\nna) admissible estimations without label shift, but with an
increased \(\lambda\) the performance further improves. For \gtbs and \nns, the
largest ratio of domain pruning is achieved with an adjustment factors of 0.4
and 0.3, whereas for the asymmetric versions \(\lambda = 0.2\) is the best
value. The best adjustment for the linear model is 0.6, and 0.4 for SVM. The
asymmetric estimators already overestimate the actual objective value and
therefore only need a small \(\lambda\).

In conclusion, applying label shift is beneficial to increase the number of
admissible estimations and to control the effectiveness of boundary estimation.
When choosing \(\lambda\) for a different setting, a reasonable value is
\(\lambda \in [0.1, 0.5]\), it is therefore recommendable to start with a low
\(\lambda\) and increase only, if the number of admissible
estimations is insufficient.

\subsection{Estimation Performance}

We analyze here the capability of each model to estimate tight domain boundaries, as compared to the original domains of the COP. As evaluation metrics, the size of the estimated domain is compared to the original domain size. Furthermore, the distance between cutting boundary and optimal objective value is compared between the estimated and original domain. A closer gap between cutting bound and objective value leads to a relatively better first solution when using the estimations and is therefore of practical interest.
Table \ref{tab:prediction} shows the estimation performance per problem and estimator.
The results show that asymmetric models are able to estimate closer boundaries than symmetric models. For each model, the estimation performance is consistent over all problems. 

\begin{table}[th]
  \centering
  \small
  \caption{Reduction in objective domain through estimated boundaries (in \%). 
  \emph{Gap}: Domain size between cutting boundary and optimum ($(1 - (|\estzub - \zopt|/|\ub{\z} - \zopt|)) * 100$).
  \emph{Size}: Ratio between new and initial domain size ($(1 - (|\estzub - \estzlb|/|\ub{\z} - \lb{\z}|)) * 100$).
  Cells show the median and the median absolute deviation (MAD): No superscript indicator \(\leq
  5 \leq {}^{+} \leq 10 < {}^{*} \leq 20 < {}^{**} \leq 30 < {}^{***}\).\\\label{tab:prediction}%
}
\begin{tabularx}{\textwidth}{lRR|RR|RR|RR|RR|RR}
\toprule
{} & \multicolumn{2}{c}{GTB$_a$} & \multicolumn{2}{c}{GTB$_s$} & \multicolumn{2}{c}{LR} & \multicolumn{2}{c}{NN$_a$} & \multicolumn{2}{c}{NN$_s$} & \multicolumn{2}{c}{SVM} \\
{} &    Gap &       Size &    Gap &      Size &     Gap &      Size &      Gap &       Size &    Gap &       Size &    Gap &      Size \\
\midrule
Bin Packing   &   68 &   65 &   60 &  58 &    48 &  48 &    \textbf{78}\textsuperscript{*} &  \textbf{68}\textsuperscript{*} &   50 &   48 &   15 &  18 \\
Cutting Stock &  \textbf{64}\textsuperscript{*} &   66 &  58\textsuperscript{*} &  59 &   48\textsuperscript{*} &  49 &  41\textsuperscript{***} &  \textbf{71}\textsuperscript{+} &  48\textsuperscript{*} &   49 &  29\textsuperscript{*} &  17 \\
Jobshop   &   69 &   69 &   60 &  60 &    50 &  50 & \textbf{87} &   \textbf{81} &   50 &   48 &   19 &  20 \\
MRCPSP    &   64 &   61 &   60 &  59 &    49 &  49 & \textbf{80} &   \textbf{76} &   49 &   49 &   13 &  19 \\
Open Stacks   &  \textbf{64}\textsuperscript{*} &  \textbf{60}\textsuperscript{+} &  59\textsuperscript{*} &  53 &  43\textsuperscript{**} &  43 &   56\textsuperscript{**} &  33\textsuperscript{*} &  47\textsuperscript{*} &  42\textsuperscript{+} &  15\textsuperscript{+} &  15 \\
RCPSP     &   65 &   64 &   60 &  60 &    50 &  50 &    \textbf{80}\textsuperscript{+} &  \textbf{76}\textsuperscript{+} &   50 &   50 &   13 &  20 \\
VRP   &   70 &   70 &   60 &  60 &    50 &  50 & \textbf{89} &   \textbf{88} &   50 &   50 &    0 &   0 \\
\bottomrule
\end{tabularx}
\end{table}

First, we look at the share of admissible estimations. Most models achieve $100$ \% admissible estimations in all problems. Exceptions exist for Cutting Stock (\gtba , \gtbs, LR: 91 \%, SVM: 50 \%) and RCPSP (\nns, SVM: 83 \%, all other models: $\ge 98 \%$). In general, \nna has the highest number of admissible estimations, followed by \gtba. 
The largest reduction is achieved by \nna, 
making it the overall best performing model.
\gtba is also capable to consistently reduce the domain size by over $60$ \%, but not as much as \nna.
Cutting Stock and Open Stacks are difficult problems for most models, as indicated by the deviations in the results.
LR and \nns reduce the domain size by approximately 50 \%, when the label shift adjustment factor $\lambda$ is $0.5$, as selected in the previous experiment.

Conclusively, these results show that \method has an excellent ability to derive general estimation models from the extracted instance features. The estimators reduce substantially the domains and provide tight boundaries.

\subsection{Effect on Solver Performance}
Our final experiment investigates the effect of objective boundaries on the actual solver performance, as described in Section~\ref{sec:method_cp}.
This includes both estimated boundaries as well as fixed boundaries calculated from the known best solution and the first solution found by the solver without boundary constraints.
The goal for this combination of estimated and fixed boundaries is to understand how helpful objective boundaries actually are and also how well boundary estimation provides useful boundaries.
By setting the fixed boundary in relation to the first found solution, we enforce an actual, non-trivial reduction in the solution space for the solver.

The setup for the experiments is as follows.
For each COP, 30 instances are randomly selected.
Each instance is solved using four distinct configurations, namely
\begin{enumerate}
\item the unmodified model without boundary constraints;
\item the estimations from \nna as upper and lower boundary constraints;
\item the estimations from \nna, only as an upper boundary constraint;
\item a fixed upper boundary of the middle between the optimum and the first found solution when using no boundary constraints ($\z_{first}$): $\ub{z} = \zopt + \lfloor(\z_{first} - \zopt) / 2\rfloor$.
\end{enumerate}
We selected three distinct solvers with different solving paradigms and features:
Chuffed (as distributed with MiniZinc 2.1.7) \cite[]{Chu2016}, Gecode 6.0.1 \cite[]{Schulte2018}, 
and Google OR-Tools 6.8.
These solvers represent state-of-the-art implementations, as shown by their high rankings in the MiniZinc challenges.
All runs are performed with a 4 hour timeout on a single CPU core of an Intel E5-2670 with 2.6 GHz.

Three metrics are used for evaluation (all in \,\%), each comparing a run with some boundary constraint to the unmodified run.
The \emph{Equivalent Solution Time} compares the time it takes the unmodified run to reach a solution of similar or better quality than the first found solution when using boundary constraints. It is calculated as 
$(t_{Original} - t_{Bounds})/t_{Original} * 100$.
The \emph{Quality of First} compares the quality of the first found solutions with and without boundary constraints and is calculated as $(1 - \z_{Bounds} / \z_{Original}) * 100$.
The \emph{Time to Completion}, finally, relates the times until the search is completed and the optimal solution is found. It is calculated in the same fashion as the Equivalent Solution Time.

The results are shown in Table~\ref{tab:solver}, listed per solver and problem.
We do not list the results for the Cutting Stock problem for Chuffed and OR-Tools, because none of the configurations, including the unmodified run, found a solution for more than one instance. 
Gecode found at least one solution for 26 of 30 instances, but none completed, and we include the results for the 26 instances.
For all other problems and instances all solvers and configurations found at least one solution.

\begin{table}[th]
\footnotesize
\centering
  \caption{Effect of boundaries on solver performance (in \%).
  \emph{Fixed}: Upper boundary set to middle between optimum and first found solution of unbounded run.
  \emph{Upper}: Upper boundary set to estimated boundary.
  \emph{Both}: Upper and lower boundary set to estimated boundaries.
  Results are averaged over 30 instances, lower values are better.\\\label{tab:solver}
  }
\begin{tabularx}{\textwidth}{lRRR|RRR|RRR}
\toprule
{} & \multicolumn{3}{c}{Equiv. Solution Time} & \multicolumn{3}{c}{Quality of First} & \multicolumn{3}{c}{Time to Completion} \\
{} & Fixed &  Upper &  Both & Fixed & Upper &  Both & Fixed &   Upper &    Both \\
\midrule
Bin Packing   &  -9.4 &   36.1 &  13.0 & -37.9 & -57.7 & -57.7 &  36.1 &  2140.7 &  2364.5 \\
Jobshop       & -96.5 &  -96.4 & -96.6 & -38.1 & -60.0 & -60.0 & -27.6 &   -53.6 &   -42.5 \\
MRCPSP        &   0.0 &    0.0 &   0.0 & -10.8 &  -0.4 &   0.0 &   1.2 &     0.3 &    -3.4 \\
Open Stacks   &  -1.3 &   -1.3 &  -0.9 & -24.0 & -13.2 & -13.2 &   2.0 &    -0.4 &     2.9 \\
RCPSP         &  -3.2 &  197.4 &  25.3 &  -3.3 &   0.0 &   0.0 &  -4.2 &     0.0 &    -4.2 \\
VRP           &   0.4 &    0.0 &   0.0 & -23.5 &   0.0 &   0.0 &   2.0 &     7.0 &     7.0 \\
\bottomrule
\end{tabularx}
\\
\begin{center}(a) Chuffed\end{center}
\begin{tabularx}{\textwidth}{lRRR|RRR|RRR}
\toprule
Bin Packing   &    53.5 &   0.3 &  -0.6 &  -4.7 &   0.0 &   0.0 & -10.3 &  -4.0 & -13.0 \\
Cutting Stock &  5627.0 &   7.3 & -29.5 &  -8.5 &  -5.5 &  -2.6 &   --  &   --  &   -- \\
Jobshop       &   189.3 &  -6.4 &  37.4 & -10.9 &   6.1 &   6.1 &   0.0 &   0.0 &   0.0 \\
MRCPSP        &     0.0 &   0.0 &  23.6 & -10.8 &  -0.4 &  -0.2 &   1.3 &   0.0 &   4.0 \\
Open Stacks   &     0.0 &  -1.5 &   0.0 & -24.0 & -12.8 & -12.8 &   8.9 &   6.4 &   6.8 \\
RCPSP         &   -17.2 &  56.8 & -14.4 &  -2.8 &   0.0 &   0.0 & -11.8 &  12.0 &  -9.4 \\
VRP           &     0.0 &   0.0 &   0.0 & -21.0 &   0.0 &   0.0 & -19.0 & -18.0 &  -8.0 \\
\bottomrule
\end{tabularx}
\\
\begin{center}(b) Gecode\end{center}
\begin{tabularx}{\textwidth}{lRRR|RRR|RRR}
\toprule
Bin Packing   & -22.7 &   35.0 &  39.2 & -37.4 & -57.0 & -57.2 &  104.4 &  170.0 &  172.4 \\
Jobshop       &   1.1 &    0.0 &   0.0 & -16.5 &  -0.8 &  -0.8 &    0.0 &    0.0 &    0.0 \\
MRCPSP        &  -3.2 &   -3.0 &  45.3 & -10.8 &  -0.4 &   0.0 &   -2.4 &   -2.1 &    1.2 \\
Open Stacks   &  -5.0 &   -2.6 &  -3.1 & -24.0 & -13.2 & -13.2 &    6.3 &   -1.2 &    2.3 \\
RCPSP         &   0.0 &  147.2 &  30.4 &  -3.3 &   0.0 &   0.0 &   -6.6 &   27.0 &    7.8 \\
VRP           & -95.3 &    0.0 &   0.0 & -38.2 &   0.0 &   0.0 &   32.0 &   -3.0 &   -5.0 \\
\bottomrule
\end{tabularx}
\\
\begin{center}(c) OR-Tools\end{center}
\end{table}

We obtain mixed results for the different solvers and problems, which indicates that the capability to effectively use objective boundaries is both problem- and solver-specific, and that deploying tighter constraints on the objective domain is not beneficial in every setting.
This holds true both for the boundaries determined by boundary estimation (columns \emph{Upper} and \emph{Both}) and the fixed boundary, which is known to reduce the solution space in relation to the otherwise first found solution when using no boundary constraints.
The general intuition, and also confirmed by the literature, is that in many cases a reduced solution space allows more efficient search and for several of the COPs, this is confirmed.
An interpretation for why the boundary constraints in some cases hinder effective search, compared to the unbounded case, is that the solvers can apply different techniques for domain pruning or search once an initial solution is found. 
However, when the solution space is strongly reduced finding this initial solution is difficult as the right part of the search tree is only discovered late in the process.

The best results are obtained for solving Jobshop with Chuffed, where the constraints improve both the time to find a good initial solution and the time until the search is completed.
Whether both an upper and lower boundary constraint can be useful is visible for the combination of Gecode and RCPSP.
Here, posting only the upper boundary is not beneficial for the Equivalent Solution Time, but with both upper and lower boundary Gecode is 14\,\% faster than without any boundaries.
Chuffed has a similar behaviour for RCPSP regarding Time to Completion, where the upper boundary alone has no effect, but posting both bounds reduces the total solving time by 4\,\%.

At the same time, we observe that posting both upper and lower boundaries, even though they reduce the original domain boundaries, do not always help to improve the solver performance.
Two examples are the combination of Chuffed and Jobshop or OR-Tools with Open Stacks.
In both cases does the lower boundary constraints reduce the performance compared to the behaviour with only the upper boundary, here in terms of Time to Completion.

In addition to the specific solver implementations, a reasons for the effectiveness of objective boundaries  the actual ability of the COP model to propagate the posted boundary constraints seems relevant.
However, when considering the relation between the COPs objective function formulations (see Table~\ref{tab:instances}) and effectiveness of boundary constraints, we do not observe a strong link in our results, that would clearly explain the measured results.

In conclusion, objective boundary constraints can generally improve solver performance.
Still, there are dependencies on the right combination of solver and COP model to make best use of the reduced domain.
Which combination is most effective and what the actual reasons for better performance are is yet an open question and, to the best of our knowledge, has not been clearly answered in the literature.
From the comparison with a fixed objective boundary that is known to reduce the solution space, we observe that the estimated boundaries are often competitive and provide a similar behaviour when the external requirements are met.
This makes boundary estimation a promising approach for further investigation and, once the external requirements on the combination of solver and COP model are properly identified, also practical deployment.

\section{Conclusion} \label{sec:conclusion}

Predictive ML has been shown to be very successful in supporting many important applications of CP, including algorithm configuration and selection, learning constraint models, and providing additional insights to support CP solvers.

In the first part of the paper, we presented the integration in these applications and discussed necessary components for their success, such as data curation and trained ML models.
Given the increasing interest in ML and the vast development of the field, we expect integration of predictive ML and CP to receive further attention as well.
We broadly identified three types of integration, that we expect to be most relevant for future applications and research: 
The first is to have a ML model included in the solver itself and thereby to foster either learning during infer and search or lifelong learning of a solver. In this scenario, ML helps for configuration, filtering consistency and propagators selection, labelling heuristics selection and potentially for any solver decisions.
Second, embedding ML models into the constraint model at compile time. Combining ML and CP in one model allows us to model problems that can not be solved by one of the paradigms alone. Furthermore, interaction of both paradigms can potentially enable further synergies at solving-time.
The third and final category is a loose coupling between ML and CP by having a solver- and model- external ML component that can make predictions, which can affect both the solver and the model. 
Each of these three integration types has advantages and drawbacks and different application areas for which they are the most appropriate.

In the second part of the paper, we presented one application of predictive ML for CP, namely \BE.
We introduced \method, a novel boundary estimation method for constraint optimization
problems (COP), which belongs to the third type mentioned above. 
A ML model is trained to estimate close boundaries
on the objective value of a COP instance. To avoid inadmissible misestimations,
training is performed using asymmetric loss and label shift, a technique to
automatically adjust training labels.

Boundary estimation has its strength in the combination of data-driven machine
learning and exact constraint optimization through branch-and-bound. Decoupling
these two parts enables broad adaptation to different problems and compatibility
with a wide range of solvers. Because estimator training requires a set of
sample instances, \method is suited for scenarios where a COP has to be
frequently solved with different data inputs. After additional instances have
been solved, it is then possible to retrain the model to improve the
estimations.
An experimental evaluation showed the capability to estimate admissible
boundaries and prune the objective domain with marginal overhead for the solver.
Testing the estimated boundaries on CP solvers showed that boundary estimation
is effective to support solving COPs.

The example of boundary estimation with \method motivates the potential advantage of integrating ML and constraint optimization, that we identified and discussed in the first part of this article.
At the same time, it stresses the necessity for data-efficient learning models and generic instance
representations to capture relevant problem and instance information.

\section*{Acknowledgment}
This research was supported by the Research Council of Norway through the T-Largo grant (Project No.: 274786) and the European Commission through the H2020 project AI4EU (Project No.: 825619).

\bibliographystyle{plainnat}
\bibliography{refs}
\end{document}